\documentclass[lettersize,journal]{IEEEtran}
\usepackage[dvipsnames]{xcolor}
\usepackage[accsupp]{axessibility}
\usepackage{amsmath,amsfonts}
\usepackage{algorithmic}
\usepackage{array}
\usepackage[caption=false,font=normalsize,labelfont=sf,textfont=sf]{subfig}
\usepackage{textcomp}
\usepackage{stfloats}
\usepackage{url}
\usepackage{booktabs} 
\usepackage{verbatim}
\usepackage{graphicx}
\newcommand{\Tref}[1]{Table~\ref{#1}}
\newcommand{\eref}[1]{Eq.~\eqref{#1}}

\newcommand{\fref}[1]{Fig.~\ref{#1}}

\newcommand{\sref}[1]{Sec.~\ref{#1}}

\usepackage{xcolor}
\newcounter{todos}
\AtEndDocument{\ifnum\value{todos}>0 \PackageWarning{TODOS}{There are \arabic{todos} todos left in this paper! Fix them before submitting the paper!} \fi}

\newcommand{\etal}{\textit{et al.}\xspace}

\usepackage{bm}
\newcommand{\V}[1]{\ensuremath{\bm{#1}}}

\usepackage{xspace}

\def\eg{\emph{e.g.}}
\def\ie{\emph{i.e.}}
\newcommand{\ap}{ApP\xspace}
\newcommand{\st}{StP\xspace}
\newcommand{\aps}{ApPs\xspace}
\newcommand{\sts}{StPs\xspace}
\newcommand{\real}{\mathbb{R}}
\newcommand{\model}{GaussianPlant\xspace}

\newcommand{\vsp}{\vspace{2mm}}
\newcommand{\data}[1]{\scalebox{0.90}{\sc #1}}

\def\BibTeX{{\rm B\kern-.05em{\sc i\kern-.025em b}\kern-.08em
    T\kern-.1667em\lower.7ex\hbox{E}\kern-.125emX}}
\usepackage{balance}

\begin{document}

\title{GaussianPlant: Structure-aligned Gaussian Splatting for 3D Reconstruction of Plants}

\author{Yang Yang$^{1}$
\quad
Risa Shinoda$^{1}$
\quad
Hiroaki Santo$^{1}$
\quad
\quad
Fumio Okura$^1$\\
$^1$The University of Osaka\\
}

\maketitle

\begin{abstract}
We present a method for jointly recovering the appearance and internal structure of botanical plants from multi-view images based on 3D Gaussian Splatting (3DGS). While 3DGS exhibits robust reconstruction of scene appearance for novel-view synthesis, it lacks structural representations underlying those appearances (\eg, branching patterns of plants), which limits its applicability to tasks such as plant phenotyping. To achieve both high-fidelity appearance and structural reconstruction, we introduce \model, a hierarchical 3DGS representation, which disentangles structure and appearance. Specifically, we employ structure primitives (\sts) to explicitly represent branch and leaf geometry, and appearance primitives (\aps) to the plants' appearance using 3D Gaussians. \sts represent a simplified structure of the plant, \ie, modeling branches as cylinders and leaves as disks. To accurately distinguish the branches and leaves, \st's attributes (\ie, branches or leaves) are optimized in a self-organized manner. \aps are bound to each \st to represent the appearance of branches or leaves as in conventional 3DGS. \sts and \aps are jointly optimized using a re-rendering loss on the input multi-view images, as well as the gradient flow from \ap to \st using the binding correspondence information. We conduct experiments to qualitatively evaluate the reconstruction accuracy of both appearance and structure, as well as real-world experiments to qualitatively validate the practical performance. Experiments show that the \model achieves both high-fidelity appearance reconstruction via \aps and accurate structural reconstruction via \sts, enabling the extraction of branch structure and leaf instances.
\end{abstract}

\section{Introduction}
\IEEEPARstart{R}{ecovering} the internal structure of objects from images, \ie, estimating underlying structures that support their shape and appearance, is a long-standing yet underexplored challenge in computer graphics (CG) and vision (CV). 
Among various object categories, plants pose particularly challenging cases due to their complex branching architectures and dense foliage, where substantial portions of the internal structure are occluded.
Accurately reconstructing plant branch skeletons and leaf instances is crucial for applications such as high-throughput phenotyping, growth modeling, botanical analysis, and creating CG assets.
Nevertheless, the recovery of these internal structural components of plants has not yet been addressed.

Traditional methods for reconstructing plant structures typically rely on explicit geometric models or hand-made rules. Skeletonization from 3D point clouds (\eg,~\cite{fu2020tree,jiang2021skeleton}) typically requires high-quality 3D scans and extensive manual pruning to remove wrongly generated branches. 
Procedural modeling, such as L-systems~\cite{Lindenmayer1968}, requires carefully tuned species-specific parameters, which hinders scalability across plants with vastly different morphologies.
These limitations highlight the need for methods that can directly capture plant structures from RGB observations, without relying on dense 3D supervision or species-specific priors.

\begin{figure}[t]
    \centering
    \includegraphics[width=\linewidth]{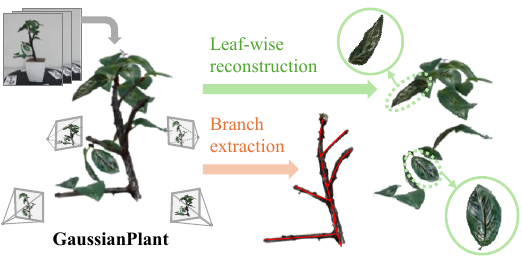}
    \caption{\textbf{\model} reconstructs both appearance and 3D structure of plants using a hierarchical 3D Gaussian Splatting (3DGS) representation, while maintaining the high capability of novel-view appearance synthesis of 3DGS. Our method enables plant-aware tasks such as branch structure extraction and leaf-wise 3D reconstruction.}
    \label{fig:teaser}
\end{figure}

To reconstruct both the structure and appearance of plants without species-specific training or manual design, we introduce \textbf{\model}, a hierarchical 3D Gaussian Splatting~(3DGS)~\cite{3dgs,dai2024high,huang20242d} representation specifically designed for plants. Leveraging the recent evolution of the family of 3DGS, the core idea of the \model is to disentangle and jointly optimize plants' structure and appearance from given multi-view images in a self-organized manner. 
Specifically, \model separates the representation into two complementary tiers. The \textbf{structure primitive (\st)} is invisible low-frequency Gaussian primitives, which are converted to cylinders or disks, representing a part of a branch or a leaf, respectively, which form the plant's structure. On the other hand, the \textbf{appearance primitive (\ap)} is a visible Gaussian primitive densely sampled and bound to the surfaces of a \st, capturing fine appearance and geometry details.

A key aspect of our method is the gradient flow between appearance and structure primitives: while \aps are optimized using photometric re-rendering losses as in standard 3DGS, \sts are refined indirectly through the motion of their bound \aps, enabling invisible structural primitives to benefit from appearance cues. This joint optimization enables \aps to progressively shape the geometry of \sts and recover internal structural components, such as branches and leaves, from multi-view RGB observations. 

Specifically, we begin by classifying ordinary 3DGS primitives into branch and leaf \sts, and attach high-frequency \aps to their surfaces. We then jointly optimize the geometry and semantic labels (\ie, branch vs.~leaf) of \sts along with the geometry and appearance parameters of \aps. The process is guided by complementary cues: an appearance-geometric cue derived from the binding relationships between \sts and \aps, and a semantic cue distilled from pretrained vision-language models. We also optimize the branches' tree graph structure, considering smoothness and connectivity, which aims to recover partially occluded branches. From the final output by \model, we can easily cluster leaf-labeled \sts to obtain leaf instances as well as extract a structural branch graph from branch \sts.

Experiments using both indoor and outdoor real-world plants show that the \model simultaneously achieves robust recovery of branch structure, leaf instances, and fine-grained appearance rendering, as shown in \fref{fig:teaser}.
This supports applications such as structural, instance-level analysis and editing, directly benefiting plant-related and CG-oriented applications.

\vsp
\noindent \textbf{Contributions.} The chief contribution of this paper is as follows:
\begin{itemize}
    \item We introduce a hierarchical 3DGS-based representation that disentangles coarse structures (\ie, \sts) and fine-grained appearances (\ie, \aps) specialized for 3D plant reconstruction. To jointly optimize them, we combine photometric supervision with geometry-aware and semantic-aware binding from an \ap to a \st, allowing for a gradient flow to update the \st's geometry and attributes (\ie, branch or leaf).
    \item We introduce a set of complementary regularizers, driven by color, semantics, and structure, that guide our model toward self-organized leaf/branch segmentation and a faithful, gap-free branching structure, without requiring manual labeling.
    \item We capture a new dataset containing real-world plants to evaluate our methods, as well as establish a benchmark for future 3D plant structure recovery and related research. 
\end{itemize}
Our implementation and the benchmark dataset will be made publicly available.


\begin{figure*}[t]
    \centering
    \includegraphics[width=\linewidth]{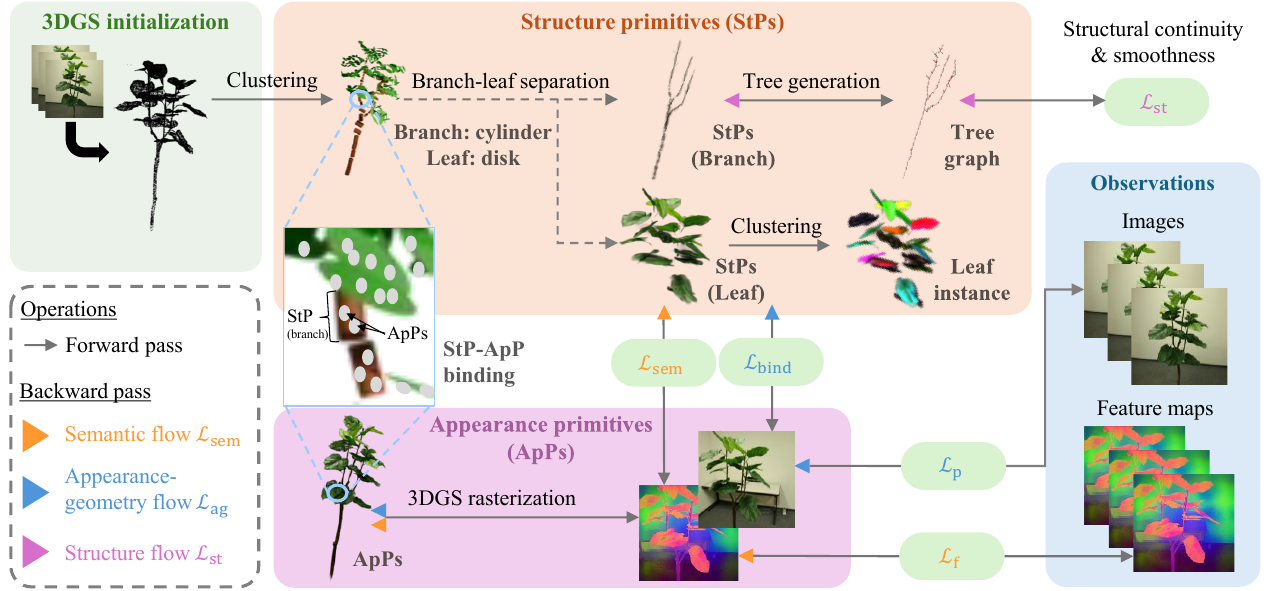}
    \caption{\textbf{Overview of \model.} Starting from a 3DGS point cloud, we cluster points into \sts and approximate them with cylinders and disks, respectively representing branches and leaves. \aps are densely located on the surface of \sts, which roles for appearance rendering similar to ordinary 3DGS. Our method jointly optimizes structure (\ie, \sts) and appearance (\ie, \aps) using binding information between \sts and \aps. 1) Appearance-geometry flow: Photometric loss computed on \aps is propagated to \sts to update both \aps' appearance and \sts' position. 2) Semantic flow: We compare the DINO features of the input images with \aps' attributes, which are also propagated to \sts to optimize the branch-leaf labels in a self-organized manner. 3) Structure flow: The structure graph extracted from \sts is evaluated to ensure the \sts are aligned to a tree graph, which optimizes the tree graph, enabling the extraction of branching structure without special post-hoc processes. }
    \label{fig:Method}
\end{figure*}

\section{Related Work}
\subsection{3D Reconstruction of Plants}
A wide variety of methods have been developed to recover the shape and structure of plants, such as leaf, branch, and root shapes from 3D observations or images~\cite{isokane2018probabilistic,Yang_2025_ICCV,liu2025treeformer,root_recon}. Classical photogrammetry pipelines use structure from motion (SfM)~\cite{schonberger2016structure} and multi-view stereo (MVS)~\cite{furukawa2015multi} to reconstruct plant-level meshes or point clouds. However, repetitive textures and severe occlusions often lead to holes or noisy geometry. Volumetric silhouettes overcome some occlusion but are limited by voxel resolution unless one resorts to costly octree optimizations~\cite{klodt2014high}. 

Recent surveys~\cite{okura20223d,li2025survey} on plant 3D reconstruction summarize a spectrum from classical multi-view geometry, highlighting the need for structural priors to recover thin branches under occlusion.
From the recovered 3D shapes, conventional skeletonization and graph‐based approaches extract branching structure as a graph via shortest‐path or minimum‐spanning‐tree (MST) algorithms (\eg,~\cite{verroust1999extracting, bucksch2010skeltre, livny2010automatic, adtree}), which assume high‐quality 3D point clouds as input (\eg, captured via LiDAR) and often break down under dense foliage. Recently, Masks-to-Skeleton~\cite{mask_to_skeleton} proposes a method to estimate a 3D tree skeleton directly from multi-view segmentation masks and a mask-guided graph optimization, which still relies on \emph{occlusion-free} 2D branch masks. Closer to our setting, Isokane~\etal~\cite{isokane2018probabilistic} infer partly hidden branch graphs from multi-view images via a generative model that reasons about occluded branches; however, they rely on a pretrained image-to-image~(\ie, leafy-to-branch) translation network that predicts per view, which is not capable of generalization across different plant species and makes cross-view consistency fragile. 

In contrast, our method leverages the self-organized optimization of a hierarchical 3DGS representation, jointly yielding a photorealistic appearance and an underlying structure, without relying on species-specific training, occlusion-free branch masks, and post-hoc skeletonization.

\subsection{3D Gaussian Splatting (3DGS)} 
3DGS~\cite{3dgs} is an emerging technique for representing and rendering 3D scenes using anisotropic Gaussian primitives. 3DGS models a scene as a collection of 3D Gaussians, each parameterized by its position, scale, orientation, opacity, and spherical harmonics for view-dependent appearance. This representation enables efficient and high-quality rendering, demonstrating significant advantages in real-time novel-view synthesis compared to NeRF-based methods (\eg, \cite{mildenhall2020nerf}).
3DGS has been widely applied in various tasks, including SLAM~\cite{matsuki2024gaussian}, text-to-3D generation~\cite{chen2024text}, human avatar modeling~\cite{hu2024gaussianavatar}, and dynamic scene reconstruction~\cite{wu20244d}. 

\subsection{3DGS for Geometry Reconstruction}
Recent efforts have explored the use of 3DGS for explicit surface extraction. SuGaR~\cite{guedon2024sugar} regularizes Gaussian locations and orientations, ensuring that they remain on and are well-distributed along surfaces. By refining an initial Poisson-reconstructed mesh, SuGaR efficiently optimizes a high-quality surface mesh in under an hour. 2D Gaussian splatting~\cite{huang20242d} takes essentially similar approach, locating 2D Gaussian primitives in a surface-aligned manner. However, while these surface-aligned approaches may successfully reconstruct the macro-scale surface of plants, they are fundamentally limited in resolving their fine-grained internal structure. Meanwhile, the primary goal of these methods is the surface shape reconstruction, which cannot directly be used for the reconstruction of the internal structure of objects (\eg, human and plant skeletons).

\subsection{3DGS for Plants} 
3DGS has been used for plant-specific applications. 
Splanting~\cite{splant2024} introduces a fast capture pipeline for plant phenotyping based on 3DGS. Wheat3DGS~\cite{zhang2025wheat3dgs} combines 3DGS with Segment Anything~\cite{kirillov2023segment} to build high-fidelity point-based reconstructions of field wheat plots and to segment hundreds of individual heads for bulk trait measurement. GrowSplat~\cite{adebola2025growsplat} constructs temporal digital twins of plants by combining 3DGS with a temporal registration. Although not focusing on reconstruction, PlantDreamer~\cite{hartley2025plantdreamer} generates realistic plants using diffusion-guided Gaussian splatting. While these methods advance plant-specific 3DGS pipelines, their primary focus is appearance reconstruction or generation. In contrast, our approach aims to go beyond appearance fidelity by explicitly recovering the plant's internal structure.

\section{\model Overview}
\model reconstructs both plant structure and appearance from multi-view RGB images using a hierarchical 3D Gaussian representation.  Here, we first briefly recap 3DGS, which is the basis of our method, followed by our \model's representation (\sref{sec:hier_gauss}) and optimization (\sref{sec:joint_opt}) methods. 

\vsp
\noindent \emph{Preliminary: 3D Gaussian splatting (3DGS).} \label{sec:prelim_3dgs}
3DGS~\cite{3dgs} uses a large set of 3D Gaussian primitives to represent a 3D scene. Each Gaussian's geometry is defined by the position of the Gaussian kernel's center $\V{\mu} \in \real^{3}$, rotation matrix $\V{R} \in \mathrm{SO}(3)$, which is internally represented by a quaternion, and scaling factors $\V{S}=\mathrm{diag}(s_1,s_2,s_3)$, where $s_i$ denotes the axis-aligned scale. The covariance matrix $\V{\Sigma}$ is defined as $\V{\Sigma}=\V{R}\V{S}\V{S}^\top\V{R}^\top$, and thus, the 3D Gaussian distribution is defined as
\begin{equation}
    G(\V{x})=\text{exp}\left(-\frac{1}{2}(\V{x}-\V{\mu})^\top\V{\Sigma}^{-1}(\V{x}-\V{\mu})\right),
\end{equation}
where $\V{x}\in\real{^3}$ is an arbitrary 3D point. To represent the appearance of Gaussian primitives, each 3D Gaussian has the color $\V{c}$ and opacity $\alpha$, where the color attribute $\V{c}$ contains the base color and spherical harmonics (SH) coefficients. 
The  color $C$ of a pixel is given by volumetric rendering along a ray:\looseness=-1
\begin{equation}
C = \sum_{i \in \mathcal{N}} c_i \alpha_i T_i ~~\text{with}~~T_i=\prod_{j=1}^{i-1}(1-\alpha_j),
\label{eq:render}
\end{equation}
where $\mathcal{N}$ is the list of 3D Gaussians whose 2D projections overlap the pixel.

\section{\model Representation: \\Hierarchical Primitives} \label{sec:hier_gauss}
We use a hierarchical representation consisting of appearance primitives (\aps) for appearance representation, and structure primitives (\sts) for representing the underlying structure, whose combination is the key component of our \model.

\sts are \emph{invisible} primitives that have a dual geometry representation: 1) 3D Gaussian for easy initialization and optimization, and 2) explicit surface to faithfully represent the structure of the plants, \ie, a cylinder if it is considered as a part of a branch and an elliptic disk if it is on a leaf. 
High-frequency \aps, represented as ordinary 3D Gaussians, are bound on the invisible \sts' surfaces and jointly optimized.

\subsection{\st: Dual-Geometry Representation}
\sts represent the coarse plant's underlying structure using invisible primitives, using both 3D Gaussian and explicit surfaces.

\begin{figure}
    \centering
    \includegraphics[width=\linewidth]{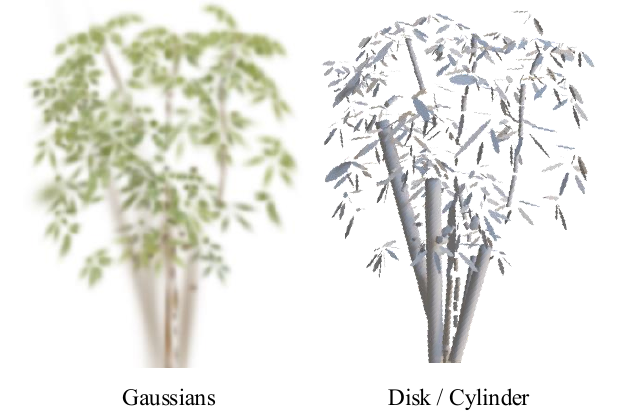}
    \caption{\textbf{Dual-geometry representation of \sts}. \sts' 3D Gaussians are converted to cylinders and disks, representing branches and leaves, respectively. Gaussian \sts' color (leaf: green, branch: brown) is just for visualization. 
    }
    \label{fig:strpr conversion}
\end{figure}

\vsp
\subsubsection{Gaussian representation}
As the same manner to the original 3DGS, a \st is represented using the center position $\V{\mu}_\text{st}\in\real^3$, rotation matrix $\V{R}_\text{st} \in \mathrm{SO}(3)$, and scale matrix $\V{S}_\text{st}=\mathrm{diag}(\V{s}_\text{st}\in\real^3)$. The \sts also equips the learnable branch-leaf probability $p_\text{st}\in(0,1)$ as their attribute, where $p_\text{st}=1$ indicates a part of a branch, and $p_\text{st}=0$ indicates a part of a leaf, respectively. 

\vsp
\subsubsection{Surface representation} 
Given the \st's parameters $\{\V{\mu}_\text{st}, \V{R}_\text{st}, \V{S}_\text{st}, p_\text{st}\}$, we convert it to explicit surface models, namely, a cylinder for branches (if $p_\text{st}\geq0.5$) and a disk for leaves (if $p_\text{st} < 0.5$), as shown in \fref{fig:strpr conversion}.

\vsp
\noindent \emph{Branches: Cylinder representation.}
When a \st is estimated as a part of a branch, \ie, $p_\text{st}\geq0.5$, the primitive is converted to a cylinder represented by the following parameters:
\begin{equation}
   \{\V{\mu_\text{cy}}=\V{\mu_\text{st}}, \:\: \V{u}_\text{cy}=\V{v}_\text{st}^1,\:\:  r_\text{cy}=s_\text{st}^2,\:\:  l_\text{cy}=3s_\text{st}^1 \},
\end{equation}
where $\V{\mu_\text{cy}}$ is the center point, $\V{u}_\text{cy}\in\real^3$ denotes the axis of the cylinder, $r_\text{cy}\in \real_+$ and $l_\text{cy}\in\real_+$ denote the radius and length, respectively. 

\vsp
\noindent \emph{Leaves: Disk representation.}
For a \st is leaf-like, \ie, $p_\text{st}<0.5$, we define the elliptic disk using a set of following surface parameters: 
\begin{equation}
       \{\V{\mu_\text{di}}=\V{\mu_\text{st}},\:\: \V{n}_\text{di}=\V{v}_\text{st}^3,\:\: a_\text{di}=2s_k^1,\:\: b_\text{di}=s_\text{st}^2\},
\end{equation}
where $\V{\mu_\text{di}}$ denotes the center of the disk, $\V{n}_\text{di}\in \real^3$ denotes the surface normal, $a\in\real_+$ and $b\in\real_+$ denote the major radius and minor radius of the disk, respectively.

\subsection{\ap: Appearance and Feature Gaussian Representation}
An \ap inherits the attributes of the original 3DGS, where its geometry is represented using the center position $\V{\mu}_\text{ap}\in\real^3$, rotation matrix $\V{R}_\text{ap} \in \mathrm{SO}(3)$, and scale matrix $\V{S}_\text{ap}=\mathrm{diag}(\V{s}_\text{ap}\in\real^3)$. For appearance representation, an \ap has its color coefficients $\V{c}_\text{ap}$ and opacity $o_\text{ap}$ in the same manner as the original 3DGS. 
An \ap is bound to a corresponding \st, where we do not explicitly denote the correspondence function for simplicity. To the optimization of branch-leaf label of corresponding \st, we also introduce a learnable semantic feature $\V{f}_\text{ap}\in\real^{D}$ for each \ap, which aggregates the image features from the observation.

\subsection{Initialization of Primitives}
We initialize \sts from an ordinary 3DGS point cloud generated from multi-view images. We first group pre-optimized 3DGS points into $k$ groups using $k$-means clustering, and then perform principal component analysis~(PCA) for each cluster to determine the Gaussian axis.
Let the center position of a cluster as $\V{\mu}_\text{st}\in\real^3$, the eigenvalues as $\left\{{\lambda}_\text{st}^1,\lambda_\text{st}^2,\lambda_\text{st}^3 \mid \left(\lambda_\text{st}^1\geq\lambda_\text{st}^2\geq\lambda_\text{st}^3\right)\right\}$, and corresponding eigenvectors as $\left\{\V{v}_\text{st}^1,\V{v}_\text{st}^2,\V{v}_\text{st}^3\right\}$.
The rotation $\V{R}_\text{st}\in\real^{3\times3}$ and scale matrices $\V{S}_\text{st}\in \mathrm{Diag}(3, \mathbb{R})$ of the corresponding \st are defined as
\begin{align}
\V{R}_\text{st} &= \begin{bmatrix} \V{v}_\text{st}^1 & \V{v}_\text{st}^2 & \V{v}_\text{st}^3 \end{bmatrix},\\
\V{S}_\text{st} &= \mathrm{diag}\left(s_\text{st}^1,s_\text{st}^2,s_\text{st}^3\right) = \alpha_\text{st}  \mathrm{diag}\left(\sqrt{\lambda_\text{st}^1}, \sqrt{\lambda_\text{st}^2}, \sqrt{\lambda_\text{st}^3}\right).
\end{align}
We initialize the branch-leaf probability $p_\text{st}$ from simple geometric cues: for each initial cluster, we measure the principal components of its points and measure the anisotropy between the dominant and secondary scales, and the third scale, which suggests the thickness. 
Clusters that are highly anisotropic and thin are treated as branch-like, whereas more planar clusters are treated as leaf-like. 
We then initialize $p_\text{st}$ to $0.6$ for branch-like primitives and $0.4$ for leaf-like ones, and let the subsequent optimization refine these probabilities. 

\aps are densely sampled on \st's explicit surfaces. The position and attributes on \aps are further optimized jointly with \sts. 
Letting the sampled point as $\V{x}\in\real^3$ and the surface normal of the \st at the point $\V{x}$ as $\V{n}\in \real^3$, the center point of the \ap is set to $\V{\mu_\text{ap}}=\V{x}$. The rotation $\V{R}_\text{ap}\in\real^{3\times3}$ is computed to align the $z$-axis with the corresponding surface normal $\V{n}$. Inspired by 2D Gaussian splatting~\cite{huang20242d}, we initialize $\V{S}_\text{ap}=\mathrm{diag}(s_x,s_y,0)$ representing a flat surface.

\section{\model Optimization} \label{sec:joint_opt} 
Our model is trained end-to-end by jointly optimizing the \sts and \aps. As summarized in \fref{fig:Method}, we organize the objective into three main flows and several regularizers:
an appearance-geometry flow  $\mathcal{L}_\text{ag}$ that couples the photometric supervision with the geometry binding between \aps and \sts, 
a semantic flow  $\mathcal{L}_\text{sem}$ that aligns \sts' semantics with \aps', 
and a structure flow $\mathcal{L}_\text{st}$ that regularizes the explicit branch graph. 
We also introduce several regularizations $\mathcal{L}_\text{reg}$. Formally, the overall objective is written as
\begin{equation}
    \mathcal{L}
    = \lambda_{\text{ag}}\mathcal{L}_{\text{ag}}
    + \lambda_{\text{sem}}\mathcal{L}_{\text{sem}}
    + \lambda_{\text{st}}\mathcal{L}_{\text{st}}
    + \lambda_{\text{reg}}\mathcal{L}_{\text{reg}},
    \label{eq:overall_loss}
\end{equation}
where $\lambda_{\text{ag}},\lambda_{\text{sem}},\lambda_{\text{st}},\lambda_{\text{reg}}$ balance their contributions.

\subsection {Appearance-Geometry Flow $\mathcal{L}_\mathrm{ag}$} 
\begin{figure}[t]
    \centering
    \includegraphics[width=\linewidth]{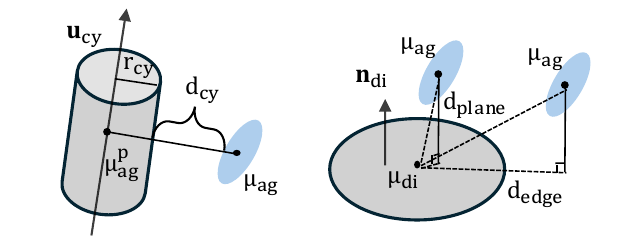}
    \caption{\textbf{Illustration of the binding loss} $\mathcal{L}_\text{bind}$. Left: the distance definition from \ap to a cylinder (\ie, branch) $d_\text{cy}$. Right: the distance definition from \ap to a disk (\ie, leaf) $d_\text{di}$, which is defined as $d_\text{plane}$ or $\sqrt{d_{\text{plane}}^{2}+d_{\text{edge}}^{2}}$~(\ie, the diagonal length in the figure). The binding loss $\mathcal{L}_\text{bind}$ is defined as a weighted sum of these two distances, $d_\text{cy}$ and $d_\text{di}$.}
    \label{fig:binding_loss}
\end{figure}

Appearance-geometry flow contain photometric loss $\mathcal{L}_\text{p}$ and binding loss $\mathcal{L}_\text{bind}$.
For photometric loss, we use the same setting as in 3DGS that combines an L1 term $\mathcal{L}_\text{L1}$ and an SSIM term $\mathcal{L}_{\text{ssim}}$ to minimize the color difference between the rendered image $\V{\hat{I}_{\text{ag}}}$ using \aps and its corresponding input image $\V{I}$:
\begin{equation}
    \mathcal{L}_{\text{p}}=  \lambda_{\text{L1}} \mathcal{L}_\text{L1}(\V{\hat{I}_{\text{ag}}}, \V{I}) +  \lambda_{\text{ssim}} \mathcal{L}_{\text{ssim}}(\V{\hat{I}_{\text{ag}}},\V{I}).
\end{equation} 
Then, binding loss $\mathcal{L}_\text{bind}$ serves as a flexible structural constraint, allowing \aps to loosely bind to a \st rather than enforcing a rigid attachment. This loose binding preserves structural consistency while still permitting local adaptation, \ie, \aps can adjust to finer appearance details while staying aligned with the overall structure. When significant structural changes occur (\eg, branch splitting), the binding loss naturally increases, signaling the need for the \st to refine and densify.
Specifically, we measure the Euclidean distance between the \ap's center and the closest point $\V{p}$ on its corresponding \st surface. For each \ap centered at $\V{\mu}_\text{ag}\in\real^3$, we define two distance functions\footnote{Hereafter, we only deal with the pairs of corresponding \ap and \st in the equations. We omit the notation of the correspondence function for simplicity.}, as illustrated in \fref{fig:binding_loss}.

Letting the corresponding cylinder's position, main axis, and radius as $\V{\mu}_\text{cy}$, $\V{u}_\text{cy}$, and $r_\text{cy}$, respectively, the \ap-to-cylinder distance $d_\text{cy}$ is given by 
\begin{equation}
  d_\text{cy} =  \max(0,\|\V{\mu}_\text{ap}-\V{\mu}_\text{ap}^\text{p}\|_2-r_\text{cy}),
\end{equation}
where $\V{\mu}_\text{ap}^\text{p}=\V{\mu}_\text{cy} + \bigl[(\V{\mu}_\text{ap} -\V{\mu}_\text{cy})\!\cdot\!\V{u}_\text{cy} \bigr]\,\V{u}_\text{cy}$ is the projection of $\V{\mu}_\text{ap}$ on the cylinder main axis $\V{u}_\text{cy}$. 

Similarly, the \ap-to-disk distance $d_\text{di}$ is
\begin{equation}
d_{\text{di}}=
\begin{cases}
d_{\text{plane}}, &
\rho\le 1,\\[6pt]
\sqrt{d_{\text{plane}}^{2}+d_{\text{edge}}^{2}}, &
\rho>1,
\end{cases}
\end{equation}
where $d_{\text{plane}}$, $d_{\text{edge}}$ refer to the perpendicular distance to the disk's plane and the planar distance from the projected point to the ellipse boundary, defined as
\begin{align}
d_{\text{plane}} &= |(\V{\mu}_\text{ap}-\V{\mu}_\text{di})\V{n}_\text{di}|,\\
d_{\text{edge}} &= \bigl\|\,[x,y]^\top\;-\left[\tfrac{a_k\,x}{\rho},\,\tfrac{b_\text{di}\,y}{\rho}\right]^\top\bigr\|_2,\\
\rho &= \sqrt{\bigl(\tfrac{x}{a_\text{di}}\bigr)^{2}+\bigl(\tfrac{y}{b_\text{di}}\bigr)^{2}},\
\end{align}
where $a_\text{di},b_\text{di}$ are the axes of the ellipse, $\rho$ measures the normalized radial distance within the ellipse, and $[x,y]^\top$ is the in‐plane coordinates of the $\V{\mu}_\text{ag}$ that are obtained by projecting onto an orthonormal basis $\V{e}_1,\V{e}_2$ of the disk's plane:
\begin{equation}
\begin{bmatrix}x\\[1pt]y\end{bmatrix}
=
\begin{bmatrix}
(\V{\mu}_\text{ap}-\V{\mu}_\text{di})\V{e}_{1}\\
(\V{\mu}_\text{ap}-\V{\mu}_\text{di})\V{e}_{2}
\end{bmatrix}.
\end{equation}

A non-trivial issue here is \emph{leaf-branch ambiguity} because the initial clustering cannot accurately determine whether a \st represents a branch or a leaf. As mentioned above, every \st is assigned with a class probability $p_\text{st}$, and we fold this uncertainty into the loss as
\begin{equation}
        \mathcal{L}_\text{bind}=\frac{1}{|{\mathcal{A}}|}\sum_{\mathcal{A}} p_\text{st}d_\text{cy} + \frac{1}{|{\mathcal{A}}|}\sum_{\mathcal{A}} (1-p_\text{st})d_\text{di}, 
\end{equation}
where ${\mathcal{A}}$ is the entire set of \aps, and $|{\mathcal{A}}|$ is the total number of \aps. The probability $p_\text{st}$ acts as a soft mask if an \ap is likely to belong to a branch \st (\ie, $p_\text{st}\approx1$), only the cylinder term matters, and vice versa for a leaf \st.
Formally, $\mathcal{L}_\text{ag}$ can be written as 
\begin{equation}
    \mathcal{L}_\text{ag} = \lambda_\text{p}\mathcal{L}_\text{p} + \lambda_\text{bind}\mathcal{L}_\text{bind},
\end{equation}
where $\lambda_\text{p}$ and $\lambda_\text{bind}$ balance the contributions of the photometric term and the geometry binding term, respectively.

\subsection{Semantic Flow $\mathcal{L}_\mathrm{sem}$}
While the geometry-based probability $p_{\text{st}}$ captures shape regularities, it may be ambiguous in regions where slender stems appear planar or overlapping leaves exhibit branch-like geometry. To enhance the discriminability of \sts, we thus incorporate semantic priors distilled from DINOv3~\cite{simeoni2025dinov3} during the optimization.  

Specifically, we assign each \ap a learnable semantic feature vector $\V{f}_\text{ap} \in \real^D$. When projecting Gaussians into the image plane, the rendered semantic feature of a pixel is computed analogously to the volumetric color compositing in \eref{eq:render}:
\begin{equation}
F_s = \sum_{i \in \mathcal{N}} \V{f}_i \alpha_i T_i. 
\end{equation}
To supervise the rendered semantic map, we first construct a ground-truth feature map $\hat{F}_s $ by extracting the DINOv3 feature map from the input images, upsampling them with a pre-trained upsampling model~\cite{couairon2025jafar}. 
Considering the computational and memory cost during rendering, we further down-project them via PCA, which also stabilizes optimization. In practice, we find that a moderate PCA dimension $D=128$ preserves the essential discriminative power of the original features while significantly reducing redundancy. Subsequently, we minimize the pixel-wise difference between the rendered semantic map and the ground-truth:
\begin{equation}
\mathcal{L}_\text{f} = \|F_s - \hat{F}_s\|_1.
\end{equation}
We then aggregate the semantic feature for each \st by pooling the features of its bound \aps:
\begin{equation}
 \V{f}_{\text{st}}=\frac{1}{|\mathcal{N(\text{st})}|}\sum_{j \in \mathcal{N}(\text{st})}\V{f}_\text{ap}^j.
\end{equation}
Then, let $\V{f}_{t} \in \real^{D}$ be DINOv3 text features projected into the same PCA space, we obtain a semantic likelihood for `leaf' and `branch' via cosine similarity as
\begin{equation}
 \pi_c=\frac{\text{exp}(\text{cos}(\V{f}_{\text{st}},\V{f}_t^c)/\tau)}{\sum_{c'\in \{ \text{branch},\text{leaf}\}}\text{exp}(\text{cos}(\V{f}_\text{st},\V{f}_t^{c'})/\tau)} .
\end{equation}
This likelihood reflects how likely a \st semantically resembles a leaf or a branch in the vision–language embedding space.
Similar to the binding loss, which leverages the same learnable probability $p_{\text{st}}$ to weight geometric distances between cylinder and disk primitives, we also use $p_{\text{st}}$ to bridge the semantic guidance by using a  cross-entropy objective
\begin{equation}
\label{eq:sem-guide}
\mathcal{L}_{\text{sem}}
= -\Big[(1-p_{\text{st}})\log \pi_{\text{leaf}}
+ p_{\text{st}}\log \pi_{\text{branch}}\Big].
\end{equation}
By optimizing both $\mathcal{L}_{\text{bind}}$ and $\mathcal{L}_{\text{sem}}$ with respect to the same $p_{\text{st}}$, the model jointly enforces geometry- and semantics-driven supervision on the \sts that ensures geometry-semantic consistency.

\subsection{Structure Flow $\mathcal{L}_{\mathrm{st}}$} 
The structure flow encourages the recovered branch primitives to form a coherent, tree-like structure with locally smooth geometry.
Here we first extract an explicit graph as the plant-structure representation, where endpoints of \sts are graph nodes and candidate connections are graph edges.
For each branch \st\ (modeled as a cylinder-like primitive), we take its center $\V{\mu}\in\mathbb{R}^3$, principal axis $\V{u}_{\text{cy}}\in\mathbb{R}^3$ ($\|\V{u}_{\text{cy}}\|_2=1$), and longitudinal scale $s_{\text{st}}^{1}>0$ to define its two endpoints
\begin{equation}
    \V{p}^{\text{top/bottom}}=\V{\mu} \pm s_{\text{st}}^{1}\V{u}_{\text{cy}}.
    \label{eq:endpoints}
\end{equation}
Each primitive thus yields one inner edge connecting its two endpoints; we denote the set of all such inner edges by $E_{\text{in}}$.

\begin{figure}[t]
    \centering
    \includegraphics[width=\linewidth]{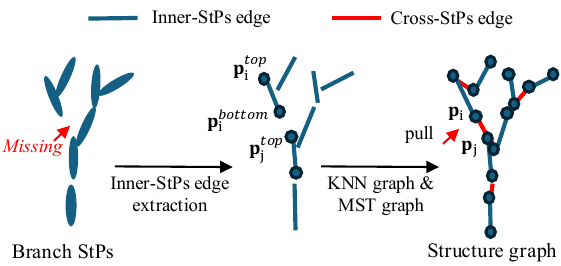}
    \caption{Structure graph built from endpoints: inner-\st\ edges (blue) and cross-\st\ MST edges (red). These edges drive the regularizers in Eq.~\eqref{eq:l_struct}.}
    \label{fig:graph_reg}
\end{figure}

To capture global tree-like topology, we form a $k$-nearest neighbors (KNN) graph over all endpoints with Euclidean weights and apply Kruskal's algorithm to extract a minimum spanning tree (MST). However, a pure Euclidean MST may introduce undesirable ``shortcut'' edges across gaps or sharp turns. Prior works often initialize tree skeletons from MST-style graphs and then prune or reweight edges to improve biological plausibility, \eg, AdTree~\cite{adtree} builds an initial MST and refines it via optimization, while earlier work by Livny et al.~\cite{livnt_tree} constructs an MST skeleton and applies global constraints to enforce a directed acyclic branch-structure graph.
Following these ideas, we therefore reweight each candidate cross-\st edge with two data-driven penalties before the final MST: Axis-consistency penalty and Void-crossing penalty.  For a cross-\st edge with  endpoints $p_i$ and $p_j$ and  their unit direction $\V{t}=\frac{(p_i-p_j)}{\|p_i-p_j\|}$, given the main axis $\V{\mu}_i$ and $\V{\mu}_j$, the axis-consistency terms $c_{ti}=|\V{t}^T\V{\mu}_i|$ and $c_{tj}=|\V{t}^T\V{\mu}_j|$ measure the collinearity between two \st axes and define the penalty as 
\begin{equation}
    \text{pen}_{\text{axis}}=1 + \gamma_{\text{tan}}(1-\frac{1}{2}(c_{ti}+c_{tj})),
\end{equation}
making oblique or misaligned connections become more expensive. To suppress`gap-crossing' shortcuts, along each cross-\st edge we sample points and count neighbors within radius $\rho$ from \aps. Edges traversing sparsely supported space are penalized as
\begin{equation}
    \text{pen}_\text{occ} = 1+\gamma_\text{occ}\frac{1}{K}\sum_k1[n_k<\theta],
\end{equation}
where $\theta$ refers to the threshold of neighbor number. The final structure-aware edge cost is 
\begin{equation}
    \omega_{ij}=d_{ij}\text{pen}_{\text{axis}}\text{pen}_{\text{occ}}.
\end{equation}
Running Kruskal's algorithm on these reweighted edges yields an MST that better follows branch directionality and avoids unsupported shortcuts. Edges in the MST that connect endpoints from different \sts are collected as cross-\st\ edges $E_{\text{cross}}$ together with $E_{\text{in}}$ they define our structure graph $G=(V, E_{\text{in}}\cup E_{\text{cross}})$ with node set $V$ the endpoints (see ~\fref{fig:graph_reg}).

Given the structural graph $G$, we encourage geometric continuity by pulling together endpoints connected by cross-\st\ edges:
\begin{equation}
    \mathcal{L}_{\text{graph}}
    = \frac{1}{|E_{\text{cross}}|}
      \sum_{(i,j)\in E_{\text{cross}}} \big\|\V{p}_i-\V{p}_j\big\|_2.
    \label{eq:l_graph}
\end{equation}
To promote local smoothness, we further apply a Laplacian penalty on the endpoint coordinates:
\begin{equation}
    \mathcal{L}_{\text{lap}}
    = \frac{1}{|V|}
      \sum_{i\in V}
      \Big\|
        \V{p}_i -
        \frac{1}{\deg(i)}
        \sum_{j\in\mathcal{N}(i)} \V{p}_j
      \Big\|_2^2,
    \label{eq:l_lap}
\end{equation}
where $\mathcal{N}(i)$ and $\deg(i)$ denote the neighbor set and degree of node $i$ in $G$.
The total structural loss is
\begin{equation}
    \mathcal{L}_{\text{st}}
    =\mathcal{L}_{\text{graph}} + \mathcal{L}_{\text{lap}}.
    \label{eq:l_struct}
\end{equation}

\subsection{Regularizers $\mathcal{L}_{\mathrm{reg}}$}
We additionally introduce several regularizers that stabilize optimization and refine the primitives: a depth loss $\mathcal{L}_\text{d}$ to better constrain geometry, an overlap loss $\mathcal{L}_\text{op}$ to discourage different \sts from collapsing onto each other, and a self-organized classification loss $\mathcal{L}_\text{cls}$ to encourage confident branch/leaf labeling.

\subsubsection{Overlap loss $\mathcal{L}_\text{op}$}
The overlap loss is introduced to enforce clear boundaries between neighboring \st, promoting a more compact and well-separated structure representation. The overlap is defined as a Gaussian-weighted Mahalanobis distance between the $i$-th \st and its nearest neighbor as
\begin{equation}
\mathcal{L}_{\text{op}} = \sum_i\sum_{j\in \mathcal{N}(i)} \text{exp}\left({-\frac{1}{2}(\V{\mu}_i-\V{\mu}_j)^\top\V{\Sigma}_j^{-1}(\V{\mu}_i-\V{\mu}_j)}\right),
\end{equation}
where $\mathcal{N}(\cdot)$ refers to the $k$-nearest neighbors, in which we set $k$ as $3$ in all experiments.

\subsubsection{Depth loss $\mathcal{L}_d$} Since monocular depth is used to guide the training in a lot of 3DGS works~\cite{3dgs, wu2024surface,li2024geogaussian}, we also incorporate this term:
\begin{equation}
    \mathcal{L}_d=\left\|\V{\hat{D}_\text{ag}}-\V{D} \right\|_2,
\end{equation}
where $\V{\hat{D}_\text{ag}}$ is the rendered depth and $\V{D}$ is the monocular depth estimated by~\cite{yang2024depth}.

\subsubsection{Classification loss $\mathcal{L}_\text{cls}$} 
To encourage each \st to confidently choose between a branch/leaf label while leveraging the color cues, we combine two terms into a classification loss between branches and leaves: 
\begin{align}
\mathcal{L}_\text{cls} &= \mathcal{L}_\text{col} + \mathcal{L}_\text{conf}, \\ 
\mathcal{L}_\text{col} &= \frac{1}{|\mathcal{A}|}\sum_{\mathcal{A}}\Bigl[p_\text{st}\,\big\|c_\text{ag} - \bar c_\text{leaf}\big\|_2^2
  +(1-p_\text{st})\,\big\|c_\text{ag} - \bar c_\text{branch}\big\|_2^2\Bigr], \\ 
\mathcal{L}_\text{conf}&=\frac{1}{|\mathcal{S}|}\sum_{\mathcal{S}}p_\text{st}\,(1-p_\text{st}),
\end{align}
where ${\mathcal{S}}$ is the entire set of \sts, and $|{\mathcal{S}}|$ is the total number of the \sts.
We use a sigmoid function to limit the range of $p_\text{st}$ in $(0,1)$, letting the probability of being a part of a branch.
Here, $c_\text{ag}$ denotes the \ap's color feature, and $\bar c_\text{leaf}$ and $\bar c_\text{branch}$ are the mean color features of the current leaf and branch sets, respectively. The first term $\mathcal{L}_\text{col}$ pulls each primitive's label toward the class whose average color it matches, while the second term $\mathcal{L}_\text{conf}$ penalizes probabilities near $0.5$ (controlled by weight $\beta$) to speed up the binarization. 

\subsection{Adaptive Density Control of Primitives} \label{sec:adapt_density}
\label{sec:density}
The densification strategy plays a crucial role in optimizing 3D Gaussians for capturing fine details while maintaining efficiency. However, the standard densification strategy in 3DGS, which relies on appearance-based gradient thresholds, is not suitable for structure extraction. We therefore introduce a new density control strategy for \sts, encompassing both densification and merging.

\subsubsection{\st densification}
The densification of \sts uses the gradient from both geometry binding loss $\mathcal{L}_\text{bind}$ and semantic binding loss $\mathcal{L}_\text{sem}$, where \sts with large binding losses are densified. This ensures that \sts remain adaptive to evolving structures without overfitting to fine-grained textures, preventing \sts from being influenced by high-frequency details. 
To maintain a compact and interpretable structure representation, we impose a hard upper bound on the total number of \sts. 
Once this bound is reached, we perform only pruning without further densification. 
During optimization, \sts with small spatial scales or low opacity are removed, as they contribute negligibly to the structural geometry. 

\begin{table}[tp]
\centering
\caption{\textbf{Quantitative comparison on novel-view synthesis} between the proposed and the baseline method averaged over all the $10$ plants. The best results are highlighted \textbf{bold}. Our method (\model) achieves better novel-view synthesis accuracies in most cases compared to the original 3DGS.}
\label{tab:performance}
\resizebox{0.8\linewidth}{!}{
\begin{tabular}{lccc}
\toprule
Method & PSNR \(\uparrow\) & SSIM \(\uparrow\) & LPIPS \(\downarrow\) \\
\midrule
3DGS~\cite{3dgs}               & 28.732  & 0.932 & 17.16\\
Ours (w/o $\mathcal{L}_{\text{bind}}$) & 28.835& 0.935 & 17.09 \\
Ours                  & \textbf{29.486} & \textbf{0.947}& \textbf{16.31}   \\
\bottomrule
\end{tabular}}
\end{table}

\subsubsection{Branch \st merging \& filtering} \label{filter}
We observe that branch \sts often fragment into many small, thin cylinders when fitting thick branches.
We thus merge nearby \sts, whose endpoints lie within a small spatial radius and whose axes align closely, computed as the cosine similarity smaller than a threshold. 
For each merged group, we collect all endpoints and compute their centroid to serve as the new cylinder center. We then perform PCA on these endpoints and set the first principal component as the new cylinder axis. By projecting each endpoint onto this axis, we obtain the minimum and maximum projection values, whose difference defines the cylinder height. Finally, we set the radius to the average of the original primitive radii, yielding a single, coherent branch cylinder represented by a \st.
Despite geometry and semantic binding providing a reasonable classification result, there still could be some misclassified \st. We add a simple radius-growth sanity check on the MST to remove noise (\ie, leaf \st). Along each parent-child edge, branch radii are expected to be non-increasing. Therefore, child \st whose radius is abnormally larger than its parent will be pruned.

\subsubsection{Leaf \st clustering for leaf instance extraction} \label{sec:clustering}
To extract leaf instances, we use a merging-based clustering approach. Note that this process is only performed once after the optimization.
Given the leaf \sts, we group those whose centers lie within a small spatial radius, as well as their normals and disk main axis, align within a predefined angular threshold.


\begin{figure*}[t]
    \centering
    \includegraphics[width=\linewidth]{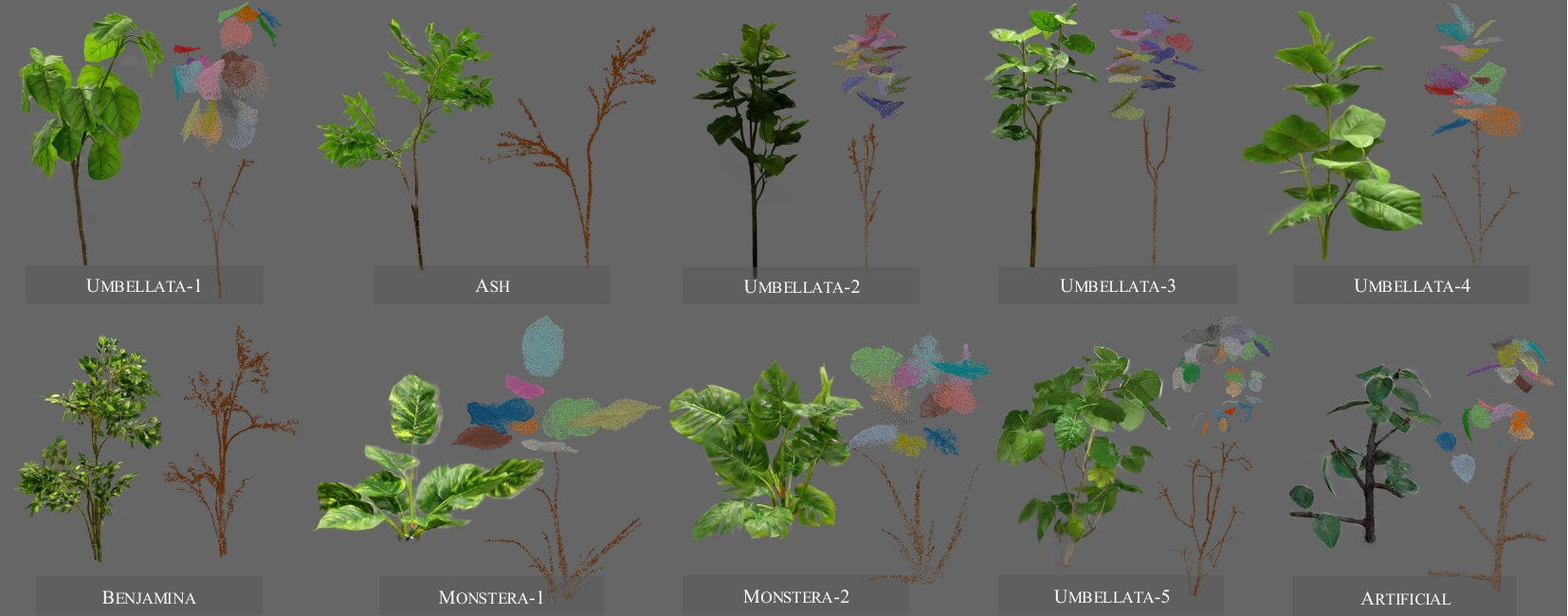}
    \caption{\textbf{\model dataset}. Ten real-world artificial plants are carefully captured in indoor environments and reconstructed with MVS. For each plant, we manually annotate the ground-truth labels of 3D branch-leaf and leaf instance segmentation (visualized on the right side of the plants' appearances).}
    \label{fig:dataset}
\end{figure*}

\section{\model Dataset \& Applications} \label{sec:apps} \label{sec:dataset} 
The task addressed by \model, \ie, 3D structure extraction of leafy plants from multi-view images, has been underexplored.
While the application field of our \model spreads beyond CV and CG, to plant science and agriculture, there are no existing datasets to assess the success of our task. We summarize here our new dataset, as well as the potential applications by \model.

\subsection{\model Dataset}
We created a new real-world benchmark dataset for 3D plant modeling and structural reconstruction, named \model dataset. The dataset contains $10$ artificial plants all captured under indoor lighting conditions, as shown in \fref{fig:dataset}. We use a variety of plant species, including those with large leaves (\eg, \data{Umbellata}) and small leaves (\eg, \data{Benjamina}), where the characteristics of shape, structure, and occlusion differ.

For each scene, we acquire more than $120$ multi-view images and reconstruct the plants using multi-view stereo\footnote{Metashape V2.2, \url{https://www.agisoftmetashape.com/}, last accessed December 4, 2025.} to create a dense and reliable point cloud. During the capture, we carefully plan viewpoints to minimize occlusions, ensuring that the reconstructed point clouds are as complete as possible.
The reconstructed point clouds are then manually cleaned to remove background and floating noise, which we use as the ground-truth shapes. 
On top of these point clouds, we annotate two types of ground-truth labels for benchmarking: (i) a binary leaf-branch segmentation and (ii) leaf instance labels.
Specifically, we import each dense point cloud into Blender\footnote{Blender, \url{http://www.blender.org/}, last accessed December 4, 2025.} and use its vertex selection tools to interactively assign labels to points. For plants with extremely dense foliage, such as \data{Ash} and \data{Benjamina}, reliable per-leaf delineation is infeasible, so we only provide binary leaf-branch labels and exclude them from the leaf instance segmentation evaluation.

\subsection{Applications}
Since our approach achieves an explicit and sparse structure representation, given the correspondence between \st and \ap, it enables a range of practical applications beyond ordinary 3D shape reconstruction and photorealistic novel-view synthesis.

\subsubsection{3D branch structure extraction}
By filtering \sts by their branch label, we obtain a compact cylinder-based skeleton and further construct a 3D graph, as introduced in \sref{sec:joint_opt}, which directly represents the connectivity and topology of the plant's branching system. Moreover, using the binding correspondence to branch \sts, we can also recover a dense branch point cloud as a subset of \aps, facilitating accurate trait measurement and structural analysis, as shown in \fref{fig:teaser}. This application can be useful for plant structural analysis in plant science and breeding, where the number of joints (\ie, tillers) and the branching pattern are important traits to evaluate the plant genotypes.

\subsubsection{Instance-wise leaf 3D reconstruction}
Leaf \sts naturally separate foliage from woody parts. We perform a simple clustering for leaf instance extraction, allowing instance-level leaf segmentation without 2D mask supervision, as shown in \fref{fig:teaser}. The 3D instance segmentation of leaves enables the direct evaluation of specific leaf traits, such as leaf area, which is directly applicable for plant phenotyping and the automation of agriculture.

\subsubsection{Plant CG asset creation and editing}
The disentangled structure and appearance representation also support interactive editing operations, such as leaf addition/removal or branch-wise deformation, enabling the creation of plant CG assets capable of dynamic simulation and specific design.


\section{Experiments}
Our primary objective is to recover the plant's structure while preserving its high-fidelity appearance. 
To this end, we  evaluate \model on four tasks:
(1) \textbf{Novel-view synthesis}, to verify that the introduction of structural primitives does not harm and can even improve image-level rendering quality;
(2) \textbf{3D branch segmentation}, to test whether our geometry-semantics-structure coupling yields cleaner leaf/branch separation than purely appearance-based, open-vocabulary segmentation on 3DGS;
(3) \textbf{Branch structure reconstruction}, to evaluate the accuracy of the recovered branch geometry beyond dense branch regions, since accurate dense branch segmentation results cannot directly demonstrate whether the recovered branches are structurally correct (\ie, a method can place points near true branches yet still produce wrong shortcuts or broken connections); and
(4) \textbf{3D leaf instance segmentation}, to examine whether the learned leaf \sts provide a more instance-friendly representation than directly clustering dense leaf \aps.

\subsection{Setup}
\subsubsection{Baselines}
For the \textbf{novel-view synthesis} task, we compare \model with the original 3DGS~\cite{3dgs}.

For the \textbf{3D branch segmentation} task, we assess the segmented 3D branch point clouds against two 3DGS-based open-vocabulary pipelines applied to the original Gaussian field.
Feature-3DGS~\cite{zhou2024feature}, which distills high-dimensional features from 2D foundation models (\eg, SAM~\cite{kirillov2023segment}, CLIP~\cite{radford2021learning}, and LSeg~\cite{li2022language}) into per-Gaussian embeddings; and 
LangSplat~\cite{qin2024langsplat}, which attaches a scale-gated affinity feature to each Gaussian and trains it with SAM-distilled, scale-aware contrastive learning.
For \model, we use the learned branch/leaf probabilities on \sts to select branch \sts and their attached dense \aps. 
For Feature-3DGS and LangSplat, we follow their open-vocabulary setting by computing the visual–text similarity between each Gaussian feature and the text prompts ``leaf'' and ``branch'', assigning the label with higher similarity, and obtaining the dense branch points. We use LSeg and CLIP as the 2D encoders for Feature-3DGS and LangSplat, respectively.

For the \textbf{branch structure reconstruction} task, to the best of our knowledge, there is no existing 3DGS-based method that outputs an explicit branch structure graph with radii comparable to our \st graph. 
We therefore evaluate only within our framework, using ablations that remove structure-related components to isolate their contribution to branch topology.

For the \textbf{leaf instance segmentation} task, we compare our \sts-based instance clustering with a classical clustering baseline, namely, DBSCAN~\cite{dbscan} applied directly on leaf \ap point clouds.

\subsubsection{Datasets}
For the quantitative evaluation, we use our \model dataset. For each plant instance, we randomly split the multi-view images of each scene into training and testing sets with a fixed ratio of $9:1$. The same train/test split is shared by all methods during evaluation.

To further test practicality and generalization, we also capture several outdoor real plants in the wild. Due to extremely dense foliage and occlusions, fine instance-level annotations are infeasible. Thus, for these scenes, we focus on the qualitative evaluation.

\begin{table*}[t]
\centering
\caption{\textbf{Quantitative comparisons of 3D branch segmentation} and ablation study. The best results are highlighted \textbf{bold}.}
\resizebox{\linewidth}{!}{
\begin{tabular}{@{\hspace{1mm}}l@{\hspace{1mm}}|cccccccccc|c}
\toprule
 & \multicolumn{11}{c}{Chamfer distance~[mm] $\downarrow$} \\
\cmidrule(lr){2-12}
Methods & \data{Umbellata-1} & \data{Ash} & \data{Umbellata-2} & \data{Umbellata-3} & \data{Umbellata-4} & \data{Benjamina} & \data{Monstera-1} & \data{Monstera-2} & \data{Umbellata-5} & \data{Artificial} & Mean \\
\midrule
Feature-3DGS~\cite{zhou2024feature} & 334.2  & 14.4   & 31.6   & 952.1 & 163.5 & 13.8 & 346.2 & 2351.0 & 178.6 & 555.2 &494.06 \\
LangSplat~\cite{li2022language}      & 171.7  & 14.5   & 7.2    & 132.8 & 18.9  & \textbf{2.0} & 124.6 & 734.2  & 82.0  & 44.5  & 133.24 \\
\midrule
Ours w/o $\mathcal{L}_\text{bind}$   & 73.8   & 8.3    & 22.6   & 75.2  & 67.5  & 33.2 & 105.4 & 116.5  & 76.8  & 67.7 & 64.7 \\
Ours w/o $\mathcal{L}_\text{sem}$    & 256.6  & 15.2   & 27.1   & 106.2 & 21.6  & 2.1  & \textbf{24.8} & 39.4   & 64.3  & 120.1 & 67.7\\
Ours w/o $\mathcal{L}_\text{op}$     & 3.5    & \textbf{2.5} & 11.2   & 96.0  & 16.5  & 8.5  & 76.4  & 23.8   & 56.4  & 45.7 & 34.1 \\
Ours w/o $\mathcal{L}_\text{cls}$    & 6.2    & 5.7    & 93.3   & 79.3  & 27.8  & 148.6 & 58.7  & 83.9   & 245.2 & 30.7 & 78.0 \\
Ours w/o $\mathcal{L}_\text{g}$      & 5.9    & 3.8    & 23.2   & \textbf{56.2} & 42.3  & 74.5 & 42.2  & 42.3   & 323.3 & 56.5 & 67.2 \\
Ours (full model)                    & \textbf{2.4} & 3.4    & \textbf{1.8} & 73.7  & \textbf{13.1} & 6.2  & 53.2  & \textbf{10.7} & \textbf{34.8} & \textbf{30.1} & \textbf{22.9}\\
\bottomrule
\end{tabular}
}
\label{tab:chamfer}
\end{table*}

\begin{figure}[t]
    \centering
    \includegraphics[width=\linewidth]{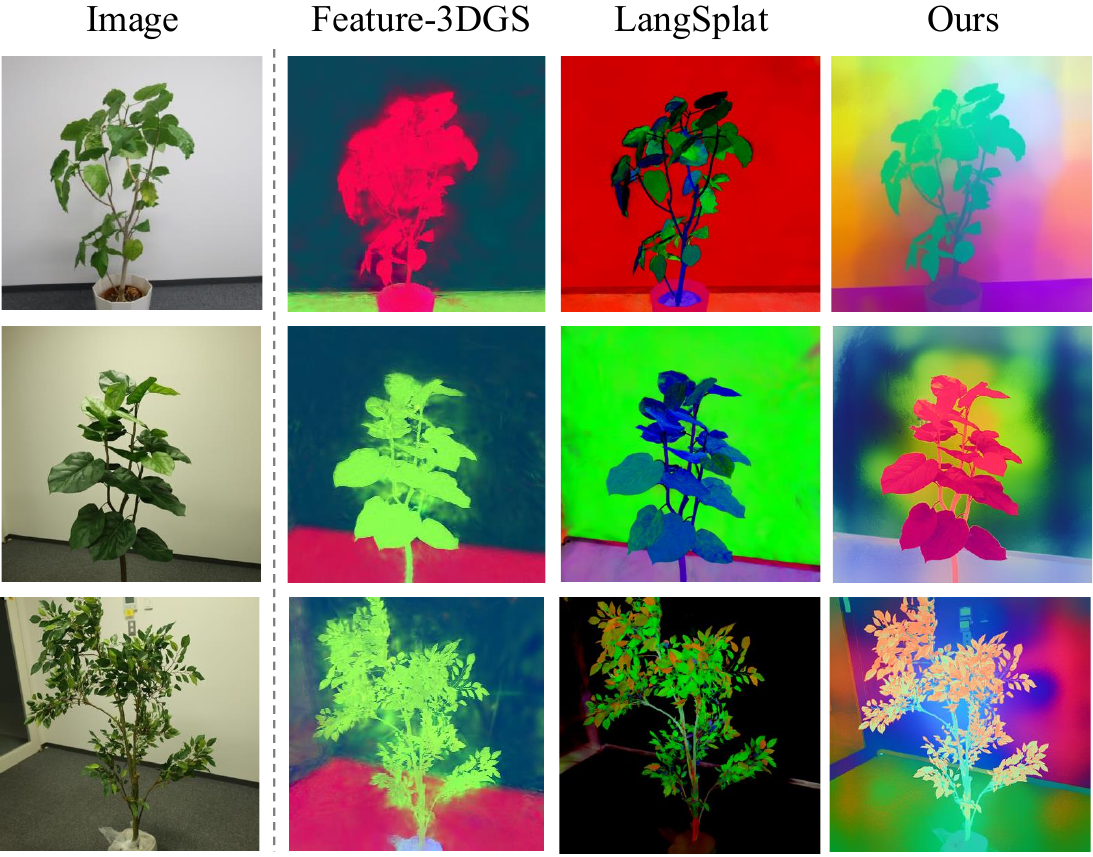}
   \caption{\textbf{Feature map visualization}. We visualize the rendered feature map on three plants under varying visibility conditions. We observe that our method gives the most accurate branch discrimination results.}
    \label{fig:feature_map}
\end{figure}

\begin{figure*}[t!]
    \centering
    \includegraphics[width=\linewidth]{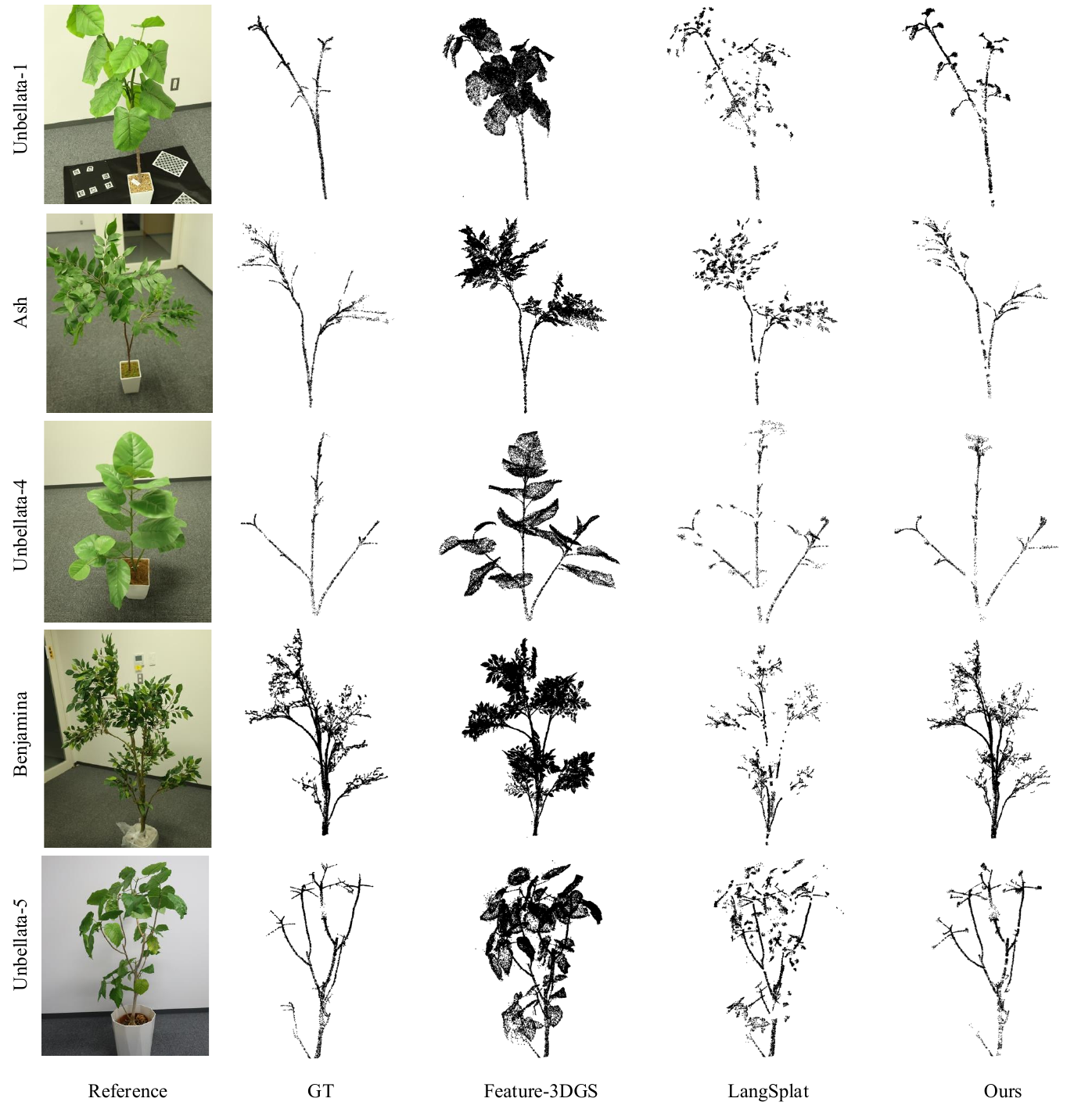}
    \caption{\textbf{Qualitative comparison on 3D branch segmentation}. From left to right: reference photo, the ground-truth branch points (GT), Feature-3DGS~\cite{zhou2024feature}, LangSplat~\cite{qin2024langsplat}, and Ours. Rows show different plants. Baselines driven purely by appearance features tend to either miss thin branchlets or include leaf clutter; our method recovers cleaner, more continuous branch scaffolds with fewer leaf artifacts.}
    \label{fig:branch_structure}
\end{figure*}

\subsubsection{Metrics} \label{sec:metric}
For the \textbf{novel‐view synthesis} task, we use the standard rendering quality metrics using PSNR, SSIM, and LPIPS for comparison.

For the \textbf{3D branch segmentation} task, we measure how accurately the segmented branch points match the ground-truth branch points. 
Given the ground-truth branch point cloud and the branch points segmented from each method, we compute the bi‐directional Chamfer Distance (CD) between the two sets.

For the \textbf{branch structure reconstruction} task, we evaluate the structural accuracy of the branch graph recovered by \sts.  
In practice, each graph edge is interpreted as a cylindrical segment whose radius is taken from the corresponding \st scale (inner-\st edges use their own radius, cross-\st edges use the mean radius of the two primitives). 
To assess the geometric accuracy of branch graphs, we densely sample points on the center of the branch \sts' cylinders, and compute the bi‐directional CD with the ground-truth branch points.

For the \textbf{leaf instance segmentation} task, we evaluate how well individual leaves are separated. We obtain dense points for each instance using the clustering label, and report the mean CD over all instances.

\subsubsection{Implementation details}
All our experiments are trained on a NVIDIA Quadro RTX 8000 GPU. Although our framework defines several objectives, we never optimize them all at once. Instead, we activate losses progressively so that each stage stabilizes the variables it is best suited for. We first pretrain Gaussians with DINOv3 distilled semantic feature using $\mathcal{L}_f$, $\mathcal{L}_d$ and $\mathcal{L}_p$ with $\lambda_p=1 ,\lambda_f=1,\lambda_d=0.1,\lambda_{\text{ssim}}=0.2,\lambda_\text{L1}=0.8$ for $30000$ iterations.

After that, we first remove the background Gaussians and initialize the \sts. Let $N_p$ denote the number of pretrained points. The initial number of clusters is set to one per 
100 points, \ie, $K=N_p/100$. A warm-up stage (first 500 iterations) is applied to initial \sts using $\mathcal{L}_p$ and $\mathcal{L}_d$ without densification to learn coarse geometry. Then, for each \st we spawn $50$ \aps, which are then optimized jointly using $\mathcal{L}_f$, $\mathcal{L}_d$, $\mathcal{L}_p$ and  $\mathcal{L}_{\text{bind}}$ with  $\lambda_p=1,\lambda_f=1,\lambda_d=0.1, \lambda_{\text{bind}}=0.1$ and follow the standard 3DGS densification strategy. 

Once \aps begin optimizing, two binding losses $\mathcal{L}_\text{sem}$ and  $\mathcal{L}_\text{bind}$ become the primary objectives for \sts, along with overlap regularizer $\mathcal{L}_\text{op}$ and classification loss $\mathcal{L}_\text{cls}$ where  $\lambda_\text{bind}=1,\lambda_\text{sem}=0.2,\lambda_\text{op}=0.05, \lambda_{\text{cls}}=0.1$.
Similar to 3DGS, we stop densification for both \aps/\sts after $15000$ iterations. Following the filtering strategy introduced in \sref{filter}, we remove likely leaf \sts and simultaneously enable the $\mathcal{L}_g$ to refine the structure with $\lambda_g=1$.

\begin{figure*}[p]
    \centering
    \includegraphics[width=\linewidth]{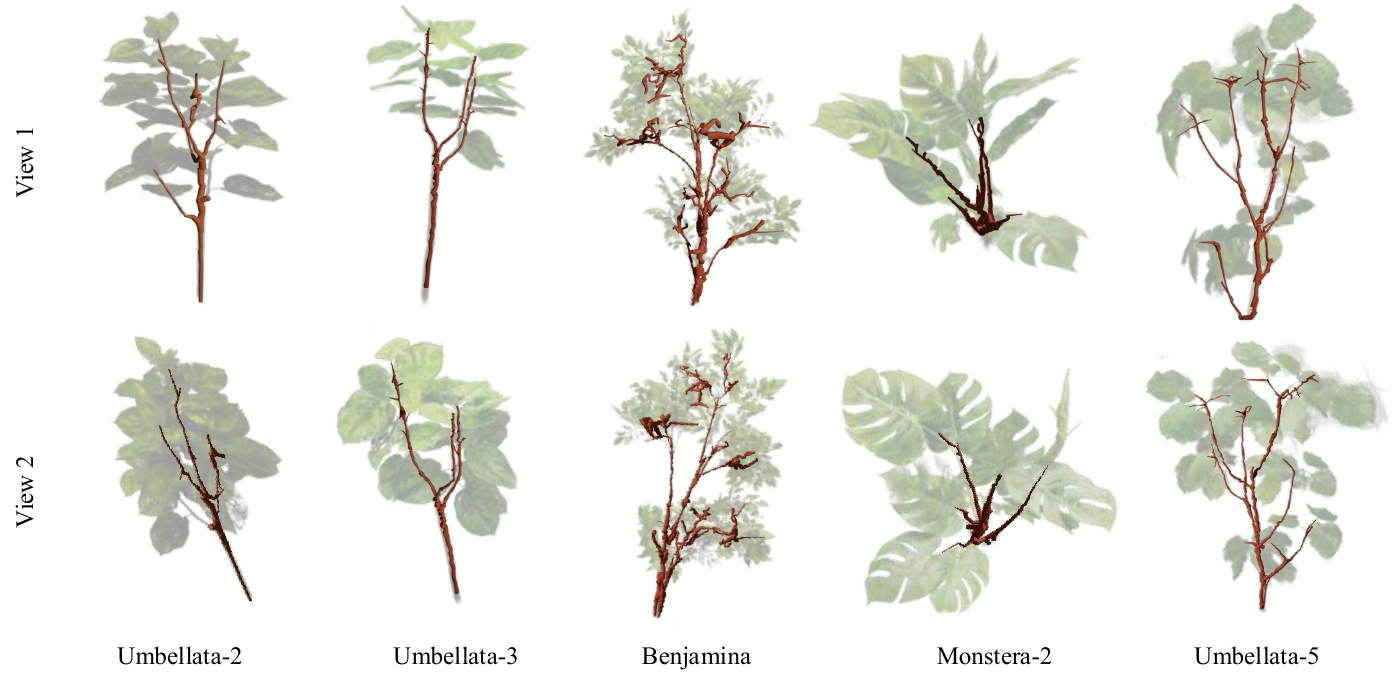}\vspace{-3mm}
    \caption{\textbf{Visual examples of branch structure reconstruction}. The front view (View 1) and the top view (View 2) show two viewpoints across plants. We render the inferred branch graph with edge volumes parameterized by the corresponding \st scales and display it in brown over the original image. Our method accurately captures the underlying branching structure, even across heavily occluded regions.}\vspace{-3mm}
    \label{fig:graph_structure}
\end{figure*}

\begin{table*}[p]
\centering
\caption{\textbf{Quantitative comparisons of branch structure reconstruction accuracy}. The best results are highlighted \textbf{bold}.}
\vspace{-3mm}
\resizebox{\linewidth}{!}{
\begin{tabular}{@{\hspace{1mm}}l@{\hspace{1mm}}|cccccccccc|c}
\toprule
 & \multicolumn{11}{c}{Structural error~[mm] $\downarrow$} \\
\cmidrule(lr){2-12}
Methods & \data{Umbellata-1} & \data{Ash} & \data{Umbellata-2} & \data{Umbellata-3} & \data{Umbellata-4} & \data{Benjamina} & \data{Monstera-1} & \data{Monstera-2} & \data{Umbellata-5} & \data{Artificial} & Mean \\
\midrule
W/o $\mathcal{L}_\text{sem}$  & 14.4 & 174.2 & 2.5  & 15.6 & 64.1 & 3.0  & \textbf{10.7} & 19.5 & 44.7 & 64.7 & 41.34 \\
W/o $\mathcal{L}_\text{bind}$ & 74.2 & 54.8  & 42.3 & 124.9& 164.5& 7.3  & 32.8        & 178.2& 64.2 & 52.2 & 79.54 \\
W/o $\mathcal{L}_\text{op}$   & 17.7 & \textbf{13.5} & 3.6  & 7.3  & 55.8 & 4.4  & 15.9        & 17.7 & 68.4 & 53.3 & 25.76 \\
W/o $\mathcal{L}_\text{cls}$  & 32.9 & 46.6  & 2.7  & \textbf{3.2}  & 86.7 & 3.6  & 24.3        & 43.2 & 73.7 & 40.1 & 35.70 \\
W/o $\mathcal{L}_\text{g}$    & 57.8 & 26.8  & 3.1  & 6.9  & 105.4& 4.7  & 41.8        & 23.3 & 96.7 & 58.7 & 42.52 \\
W/o reweighted MST            & 88.2 & 21.0  & 4.2  & 96.1 & 71.4 & 16.3 & 72.2        & 34.3 & 64.9 & 77.2 & 54.58 \\
Ours (full model)                  & \textbf{12.1} & 14.7 & \textbf{2.1} & 5.5 & \textbf{54.3} & \textbf{2.5} & 12.4 & \textbf{16.5} & \textbf{37.9} & \textbf{46.3} & \textbf{20.43} \\
\bottomrule
\end{tabular}
}
\vspace{-3mm}
\label{tab:graph}
\end{table*}

\begin{figure*}[p]
    \centering
    \includegraphics[width=\linewidth]{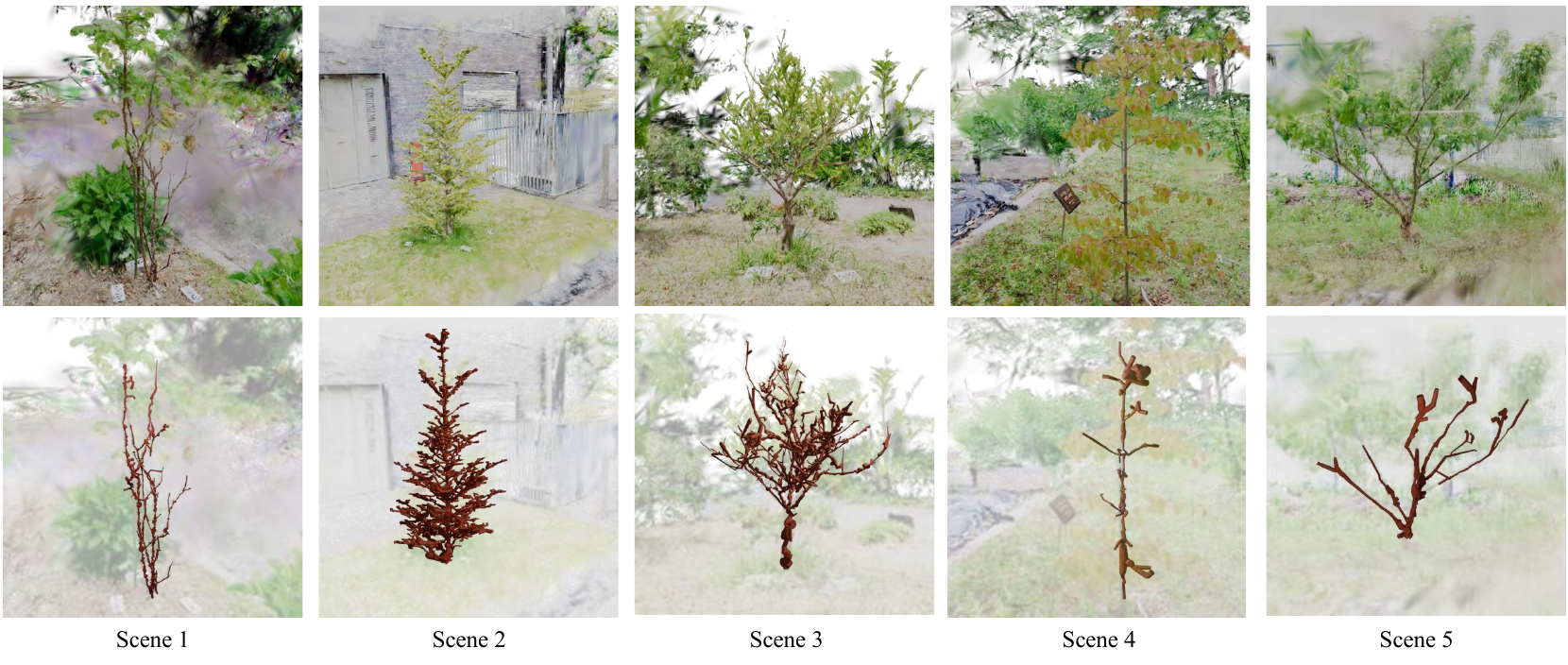}\vspace{-3mm}
    \caption{\textbf{In-the-wild applications}. Several outdoor scenes with complex geometry and strong occlusions. Top row: 3DGS renderings. Bottom row: Reconstructed branch structures. Despite clutter and heavy foliage, our method infers plausible, tree-like structures in the wild.}
    \label{fig:in-the-wild-graph}
\end{figure*}

\subsection{Results}

\subsubsection{Novel-view synthesis}
To evaluate the effectiveness of our proposed method, we conduct novel-view synthesis experiments on multiple real-world plant datasets and compare the rendering quality against the baseline 3DGS. 
As shown in \Tref{tab:performance}, our method outperforms the baseline 3DGS across all samples, demonstrating improved rendering quality. Additionally, we conduct an ablation study to analyze the impact of our proposed soft binding between \aps and \sts. Specifically, we evaluate our method without the binding loss $\mathcal{L}_{\text{bind}}$. The results show that removing the binding loss leads to a degradation in rendering quality, highlighting their importance in refining structure-aligned Gaussian splitting.

\begin{table*}[t]
\centering
\caption{\textbf{Quantitative comparisons of leaf instance segmentation accuracy}. 
We report the mean bi-directional Chamfer distance~[mm] over all leaf instances in each plant (\data{Ash} and \data{Benjamina} are excluded because their leaves are too densely packed to obtain reliable manual annotations). 
The best results are highlighted \textbf{bold}.}
\resizebox{\linewidth}{!}{
\begin{tabular}{@{\hspace{3mm}}l@{\hspace{3mm}}|cccccccc|c}
\toprule
 & \multicolumn{9}{c}{Chamfer distance~[mm] $\downarrow$ (mean over leaf instances)} \\
\cmidrule(lr){2-10}
Methods & \data{Umbellata-1} &  \data{Umbellata-2} & \data{Umbellata-3} & \data{Umbellata-4}  & \data{Monstera-1} & \data{Monstera-2} & \data{Umbellata-5} & \data{Artificial} & Mean \\
\midrule
Clustering on leaf \aps points~\cite{dbscan} & 10.1  & 1.52 & 11.1 & 1.71 & 2.64 & 4.05 & \textbf{1.31} & 6.23  & 4.83  \\
Clustering on leaf \sts (ours)     &\textbf{3.21}   &\textbf{1.05}  & \textbf{4.23} & \textbf{1.19} & \textbf{1.76} & \textbf{2.61} & 1.42  & \textbf{3.14}   & \textbf{2.33} \\
\bottomrule
\end{tabular}
}
\label{tab:leaf_instance}
\end{table*}

\begin{figure}[t]
    \centering
    \includegraphics[width=\linewidth]{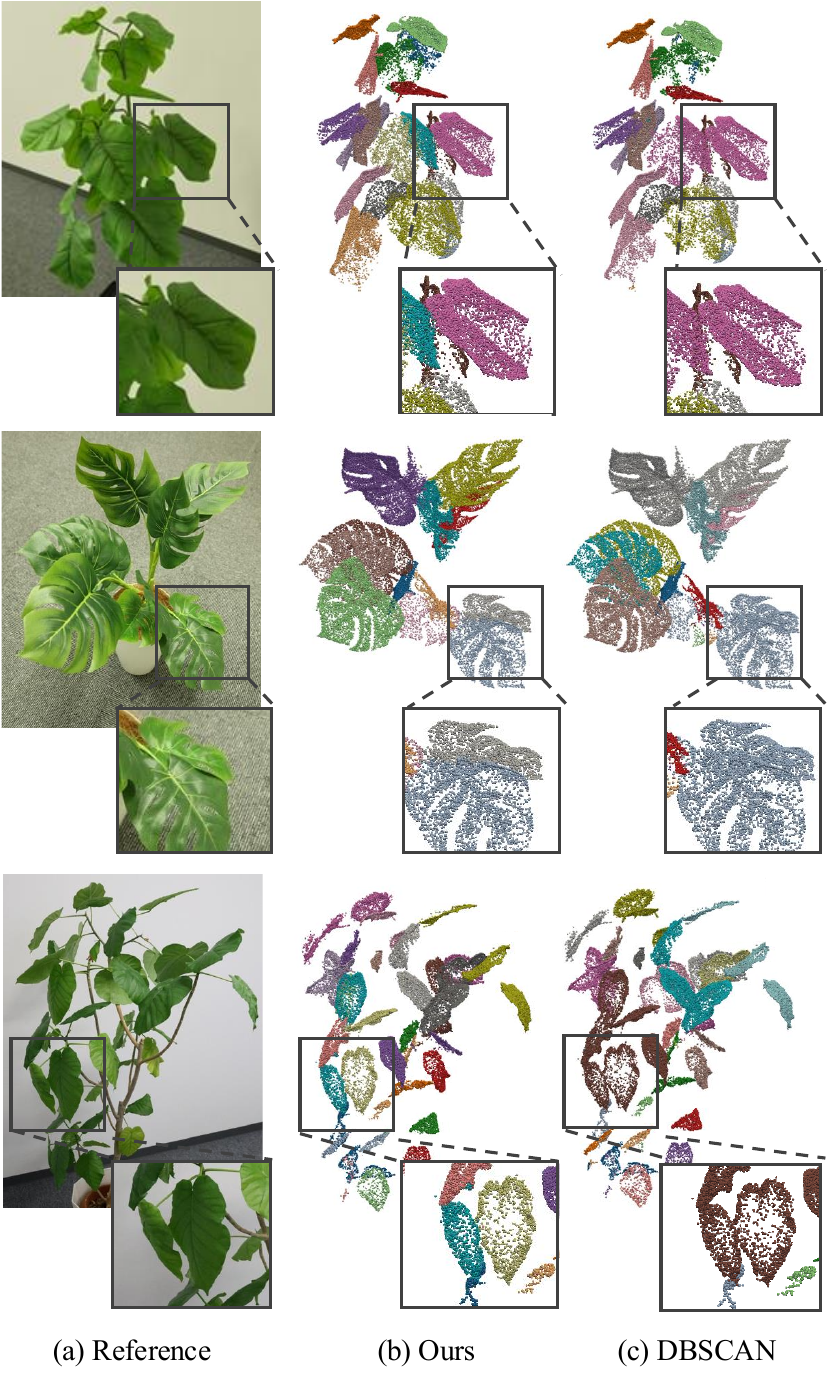} \vspace{-6mm}
    \caption{\textbf{Visualization of leaf instance segmentation results using 3DGS's point clouds.} From left to right: reference image, the clustering on leaf \sts (ours), and naive DBSCAN clustering on leaf \aps' dense points. Each predicted instance is displayed in a different color.}
    \label{fig:instance-segmentation}
\end{figure}

\begin{figure}[t]
    \centering
    \includegraphics[width=\linewidth]{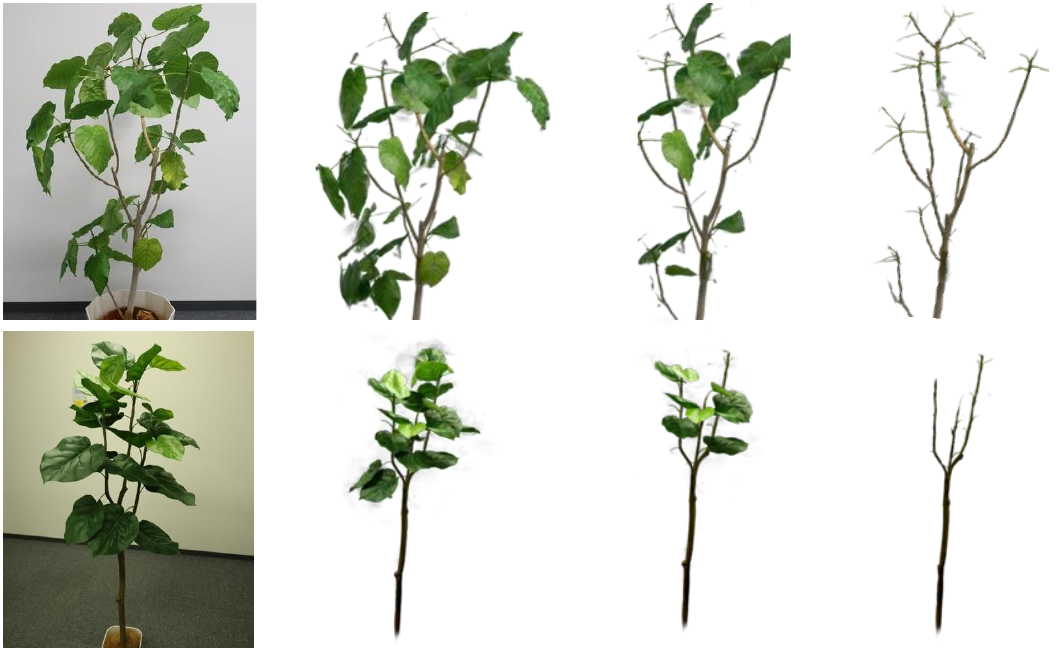}
    \caption{\textbf{Progressive leaf removal}. Instance-aware 3D leaf reconstruction can be used for practical applications. We here show an example of 3D model manipulation by removing the leaf instances' \aps according to the instance labels obtained from leaf \sts clustering. Samples from left to right show increasing removal strength, and leaving a clean branch in the last column. }
    \label{fig:leaf_removal}
\end{figure}

\subsubsection{3D branch segmentation} 
Quantitative results on 3D branch extraction are summarized in Table~\ref{tab:chamfer}, where \model achieves the lowest CD on average over all plants. Fig.~\ref{fig:feature_map} further visualizes the rendered feature maps and shows that our semantic field achieves clearer separation between leaves and branches. These observations are consistent with the visual examples shown in Fig.~\ref{fig:branch_structure}. Baselines driven purely by appearance features tend to either miss thin branchlets or be confused with leaf clutter, whereas our geometry-semantics-structure coupling recovers cleaner, more continuous branch structures. 

\subsubsection{Branch structure reconstruction}
Estimated branch graphs are visualized in Fig.~\ref{fig:graph_structure}, and the quantitative results for our ablations are summarized in \Tref{tab:graph}.  
Consistent with the evaluation on the 3D branch segmentation, our full model achieves the best accuracy. Importantly, removing structure-related components (\ie, $\mathcal{L}_\text{g}$ or the reweighted MST) noticeably degrades graph accuracy, indicating that our structure-aware optimization indeed improves the recovered branch topology. Fig.~\ref{fig:in-the-wild-graph} shows visual examples of our rendered results as well as the branch structure reconstruction for outdoor in-the-wild scenes. Our method successfully recovers plausible branching in practical scenes.

\subsubsection{3D leaf instance segmentation} 
\Tref{tab:leaf_instance} reports the quantitative comparisons over all instances, and Figs.~\ref{fig:teaser} and \ref{fig:instance-segmentation} shows the visual examples.
Our \sts-based clustering accurately segments leaf instances compared to classical clustering on the dense leaf points (recovered as \aps in our method).

The accurate leaf instance segmentation highlights applications, such as the progressive leaf removal in \fref{fig:leaf_removal}.

\section{Conclusions}
In this work, we presented \model, a hierarchical 3D Gaussian Splatting framework that jointly reconstructs plant structure and appearance from multi-view RGB imagery. Unlike conventional 3DGS methods that focus primarily on photorealistic rendering, \model explicitly disentangles coarse structural primitives (\sts) from high-frequency appearance primitives (\aps), enabling the recovery of branch topology, leaf instances, and fine-grained geometry while preserving rendering fidelity. Through geometry–appearance coupling, semantic guidance, and structure-aware regularization, our method progressively organizes 3D Gaussians into biologically meaningful representations that align with true plant architecture.

Experiments on both indoor and in-the-wild datasets demonstrate that \model achieves robust structural reconstruction, outperforming baselines in 3D branch extraction, leaf/branch segmentation, and instance-wise leaf recovery, while maintaining high-quality novel-view synthesis. The resulting structured representation further enables downstream applications such as branch graph analysis, instance-level phenotyping, and CG asset manipulation.

Looking forward, \model lays the foundation for more tightly integrated semantic-structural modeling within 3DGS. Future directions include introducing richer geometric primitives beyond cylinders and disks, improving robustness under extreme occlusion, and incorporating stronger priors or temporal cues to handle dynamic or highly complex botanical structures.

\vsp
\noindent\emph{Limitations.}
Our accuracy is bounded by (i) the fidelity of 3DGS-derived point clouds and (ii) the discriminative power of 2D foundation features projected into 3D. Dense foliage and severe occlusions remain challenging and can lead to missing or spurious branches. In addition, the current primitives (cylinders and disks) are an approximation that may be suboptimal for certain species (\eg, conifer). We plan to incorporate richer primitives (\eg, sphere and Bézier curves) and stronger priors to improve robustness under heavy occlusion and complex geometry.

\section*{Acknowledgments}
This work was supported by JSPS KAKENHI Grant Numbers JP23H05491 and JP25K03140, and JST FOREST Grant Number JPMJFR206F.


\bibliographystyle{IEEEtran}
\bibliography{main}

@String{Computer = "{IEEE} Computer" }

@String{Springer = "Springer-Verlag" }

@string{CVPR = "{Proceedings of IEEE/CVF Conference on Computer Vision and Pattern Recognition (CVPR)}"}

@string{ICCV = "{Proceedings of IEEE/CVF International Conference on Computer Vision (ICCV)}"}

@string{ICCVW = "{Proceedings of IEEE/CVF International Conference on Computer Vision Workshops (ICCVW)}"}

@string{ECCV = "{Proceedings of European Conference on Computer Vision (ECCV)}"}

@string{WACV = "{Proceedings of IEEE/CVF Winter Conference on Applications of Computer Vision (WACV)}"}

@string{ICML = "{Proceedings of International Conference on Machine Learning (ICML)}"}

@string{ICLR = "{Proceedings of International Conference on Learning Representations (ICLR)}"}

@string{SIGGRAPH = "{Proceedings of ACM SIGGRAPH Conference Papers}"}

@string{NIPS = "{Proceedings of Annual Conference on Neural Information Processing Systems (NeurIPS)}"}

@String{TOG= {ACM Transactions on Graphics (TOG)}}

@article{3dgs,
  title={{3D} gaussian splatting for real-time radiance field rendering.},
  author={Kerbl, Bernhard and Kopanas, Georgios and Leimk{\"u}hler, Thomas and Drettakis, George},
  journal=TOG,
  volume={42},
  number={4},
  pages={139--1},
  year={2023}
}

@inproceedings{guedon2024sugar,
  title={{SuGaR}: Surface-aligned gaussian splatting for efficient {3D} mesh reconstruction and high-quality mesh rendering},
  author={Gu{\'e}don, Antoine and Lepetit, Vincent},
  booktitle=CVPR,
  pages={5354--5363},
  year={2024}
}

@article{fu2020tree,
  title={Tree skeletonization for raw point cloud exploiting cylindrical shape prior},
  author={Fu, Lixian and Liu, Ji and Zhou, Jianling and Zhang, Min and Lin, Yan},
  journal={IEEE Access},
  volume={8},
  pages={27327--27341},
  year={2020},
  publisher={IEEE}
}

@article{jiang2021skeleton,
  title={Skeleton extraction from point clouds of trees with complex branches via graph contraction},
  author={Jiang, Anling and Liu, Ji and Zhou, Jianling and Zhang, Min},
  journal={The Visual Computer},
  volume={37},
  pages={2235--2251},
  year={2021},
  publisher={Springer}
}

@article{Lindenmayer1968,
  title     = {Mathematical models for cellular interactions in development II. Simple and branching filaments with two-sided inputs},
  author    = {Aristid Lindenmayer},
  journal   = {Journal of Theoretical Biology},
  volume    = {18},
  pages     = {300--315},
  year      = {1968},
  doi_       = {10.1016/0022-5193(68)90080-5},
  url_       = {https://en.wikipedia.org/wiki/L-system}
}

@inproceedings{huang20242d,
  title={{2D} gaussian splatting for geometrically accurate radiance fields},
  author={Huang, Binbin and Yu, Zehao and Chen, Anpei and Geiger, Andreas and Gao, Shenghua},
  booktitle={Proceedings of ACM SIGGRAPH},
  pages={1--11},
  year={2024}
}

@inproceedings{dai2024high,
  title={High-quality surface reconstruction using {Gaussian} surfels},
  author={Dai, Pinxuan and Xu, Jiamin and Xie, Wenxiang and Liu, Xinguo and Wang, Huamin and Xu, Weiwei},
  booktitle={Proceedings of ACM SIGGRAPH},
  pages={1--11},
  year={2024}
}

@inproceedings{wu2024surface,
  title={Surface reconstruction from {3D} gaussian splatting via local structural hints},
  author={Wu, Qianyi and Zheng, Jianmin and Cai, Jianfei},
  booktitle=ECCV,
  pages={441--458},
  year={2024}
}

@inproceedings{li2024geogaussian,
  title={Geo{G}aussian: Geometry-aware gaussian splatting for scene rendering},
  author={Li, Yanyan and Lyu, Chenyu and Di, Yan and Zhai, Guangyao and Lee, Gim Hee and Tombari, Federico},
  booktitle=ECCV,
  pages={441--457},
  year={2024}
}

@inproceedings{yang2024depth,
  title={Depth anything v2},
  author={Yang, Lihe and Kang, Bingyi and Huang, Zilong and Zhao, Zhen and Xu, Xiaogang and Feng, Jiashi and Zhao, Hengshuang},
  booktitle=NIPS,
  volume={37},
  pages={21875--21911},
  year={2024}
}

@inproceedings{zhang2025wheat3dgs,
  title={{Wheat3DGS}: In-field {3D} Reconstruction, Instance Segmentation and Phenotyping of Wheat Heads with Gaussian Splatting},
  author={Zhang, Daiwei and Gajardo, Joaquin and Medic, Tomislav and Katircioglu, Isinsu and Boss, Mike and Kirchgessner, Norbert and Walter, Achim and Roth, Lukas},
  booktitle={Proceedings of Computer Vision and Pattern Recognition (CVPR) Workshop},
  year={2025}
}

@inproceedings{splant2024,
title = {Splanting: {3D} plant capture with gaussian splatting},
author = {Ojo, Tommy and La, Thai and Morton, Andrew and Stavness, Ian},
year = {2024},
booktitle = {Proceedings of SIGGRAPH Asia Technical Communications}
}

@article{okura20223d,
  title={{3D} modeling and reconstruction of plants and trees: A cross-cutting review across computer graphics, vision, and plant phenotyping},
  author={Okura, Fumio},
  journal={Breeding Science},
  volume={72},
  number={1},
  pages={31--47},
  year={2022}
}

@inproceedings{klodt2014high,
  title={High-resolution plant shape measurements from multi-view stereo reconstruction},
  author={Klodt, Maria and Cremers, Daniel},
  booktitle=ECCV,
  year={2014},
}

@inproceedings{verroust1999extracting,
  title={Extracting skeletal curves from {3D} scattered data},
  author={Verroust, Anne and Lazarus, Francis},
  booktitle={Proceedings of International Conference on Shape Modeling and Applications},
  pages={194--201},
  year={1999}
}

@article{bucksch2010skeltre,
  title={{SkelTre}: Robust skeleton extraction from imperfect point clouds},
  author={Bucksch, Alexander and Lindenbergh, Roderik and Menenti, Massimo},
  journal={The Visual Computer},
  volume={26},
  pages={1283--1300},
  year={2010}
}

@article{livny2010automatic,
  title={Automatic reconstruction of tree skeletal structures from point clouds},
  author={Livny, Yotam and Yan, Feilong and Olson, Matt and Chen, Baoquan and Zhang, Hao and El-Sana, Jihad},
  journal=TOG,
  volume={29},
  number={6},
  pages={1--8},
  year={2010}
}

@inproceedings{matsuki2024gaussian,
  title={Gaussian splatting {SLAM}},
  author={Matsuki, Hidenobu and Murai, Riku and Kelly, Paul HJ and Davison, Andrew J},
  booktitle=CVPR,
  pages={18039--18048},
  year={2024}
}

@inproceedings{mildenhall2020nerf,
  title={{NeRF}: Representing Scenes as Neural Radiance Fields for View Synthesis},
  author={Mildenhall, Ben and Srinivasan, Pratul P and Tancik, Matthew and Barron, Jonathan T and Ramamoorthi, Ravi and Ng, Ren},
  booktitle=ECCV,
  pages={405--421},
  year={2020}
}

@inproceedings{chen2024text,
  title={{GSGEN}: Text-to-{3D} using gaussian splatting},
  author={Chen, Zilong and Wang, Feng and Wang, Yikai and Liu, Huaping},
  booktitle=CVPR,
  pages={21401--21412},
  year={2024}
}

@inproceedings{hu2024gaussianavatar,
  title={Gaussianavatar: Towards realistic human avatar modeling from a single video via animatable 3d gaussians},
  author={Hu, Liangxiao and Zhang, Hongwen and Zhang, Yuxiang and Zhou, Boyao and Liu, Boning and Zhang, Shengping and Nie, Liqiang},
  booktitle=CVPR,
  pages={634--644},
  year={2024}
}

@inproceedings{wu20244d,
  title={{4D} gaussian splatting for real-time dynamic scene rendering},
  author={Wu, Guanjun and Yi, Taoran and Fang, Jiemin and Xie, Lingxi and Zhang, Xiaopeng and Wei, Wei and Liu, Wenyu and Tian, Qi and Wang, Xinggang},
  booktitle=CVPR,
  pages={20310--20320},
  year={2024}
}

@inproceedings{schonberger2016structure,
  title={Structure-from-motion revisited},
  author={Schonberger, Johannes L and Frahm, Jan-Michael},
  booktitle=CVPR,
  pages={4104--4113},
  year={2016}
}

@article{furukawa2015multi,
  title={Multi-view stereo: A tutorial},
  author={Furukawa, Yasutaka and Hern{\'a}ndez, Carlos and others},
  journal={Foundations and Trends in Computer Graphics and Vision},
  volume={9},
  number={1-2},
  pages={1--148},
  year={2015},
  publisher={Now Publishers, Inc.}
}

@inproceedings{kirillov2023segment,
  title={Segment anything},
  author={Kirillov, Alexander and Mintun, Eric and Ravi, Nikhila and Mao, Hanzi and Rolland, Chloe and Gustafson, Laura and Xiao, Tete and Whitehead, Spencer and Berg, Alexander C and Lo, Wan-Yen and Dollar, Piotr and Girshick, Ross},
  booktitle=ICCV,
  pages={4015--4026},
  year={2023}
}

@inproceedings{dbscan,
  title={A density-based algorithm for discovering clusters in large spatial databases with noise},
  author={Ester, Martin and Kriegel, Hans-Peter and Sander, J{\"o}rg and Xu, Xiaowei},
  booktitle={Proceedings of ACM SIGKDD Conference on Knowledge Discovery and Data Mining (KDD)},
  volume={96},
  number={34},
  pages={226--231},
  year={1996}
}

@inproceedings{adebola2025growsplat,
  title={{GrowSplat}: Constructing Temporal Digital Twins of Plants with Gaussian Splats},
  author={Adebola, Simeon and Xie, Shuangyu and Kim, Chung Min and Kerr, Justin and van Marrewijk, Bart M and van Vlaardingen, Mieke and van Daalen, Tim and van Loo, EN and Rincon, Jose Luis Susa and Solowjow, Eugen and others},
  booktitle={Proceedings of IEEE International Conference on Automation Science and Engineering (CASE)},
  pages={1766--1773},
  year={2025},
}

@inproceedings{hartley2025plantdreamer,
  title={{PlantDreamer}: Achieving Realistic {3D} Plant Models with Diffusion-Guided Gaussian Splatting},
  author={Hartley, Zane KJ and Stuart, Lewis AG and French, Andrew P and Pound, Michael P},
  booktitle=ICCVW,
  year={2025}
}

@inproceedings{zhou2024feature,
  title={Feature {3DGS}: Supercharging {3D Gaussian} splatting to enable distilled feature fields},
  author={Zhou, Shijie and Chang, Haoran and Jiang, Sicheng and Fan, Zhiwen and Zhu, Zehao and Xu, Dejia and Chari, Pradyumna and You, Suya and Wang, Zhangyang and Kadambi, Achuta},
  booktitle=CVPR,
  pages={21676--21685},
  year={2024}
}

@inproceedings{qin2024langsplat,
  title={Langsplat: {3D} language gaussian splatting},
  author={Qin, Minghan and Li, Wanhua and Zhou, Jiawei and Wang, Haoqian and Pfister, Hanspeter},
  booktitle=CVPR,
  pages={20051--20060},
  year={2024}
}

@inproceedings{li2022language,
  title={Language-driven semantic segmentation},
  author={Li, Boyi and Weinberger, Kilian Q and Belongie, Serge and Koltun, Vladlen and Ranftl, Ren{\'e}},
  booktitle=ICLR,
  year={2022}
}

@inproceedings{radford2021learning,
  title={Learning transferable visual models from natural language supervision},
  author={Radford, Alec and Kim, Jong Wook and Hallacy, Chris and Ramesh, Aditya and Goh, Gabriel and Agarwal, Sandhini and Sastry, Girish and Askell, Amanda and Mishkin, Pamela and Clark, Jack and others},
  booktitle=ICML,
  pages={8748--8763},
  year={2021},
}

@article{simeoni2025dinov3,
  title={{DINOv3}},
  author={Sim{\'e}oni, Oriane and Vo, Huy V and Seitzer, Maximilian and Baldassarre, Federico and Oquab, Maxime and Jose, Cijo and Khalidov, Vasil and Szafraniec, Marc and Yi, Seungeun and Ramamonjisoa, Micha{\"e}l and others},
  journal={arXiv preprint arXiv:2508.10104},
  year={2025}
}

@inproceedings{couairon2025jafar,
  title={{JAFAR}: Jack up Any Feature at Any Resolution},
  author={Couairon, Paul and Chambon, Loick and Serrano, Louis and Haugeard, Jean-Emmanuel and Cord, Matthieu and Thome, Nicolas},
  booktitle=NIPS,
  year={2025}
}

@Article{mask_to_skeleton,
AUTHOR = {Liu, Xinpeng and Xu, Kanyu and Shinoda, Risa and Santo, Hiroaki and Okura, Fumio},
TITLE = {Masks-to-Skeleton: Multi-View Mask-Based Tree Skeleton Extraction with {3D} Gaussian Splatting},
JOURNAL = {Sensors},
VOLUME = {25},
NUMBER = {14},
YEAR = {2025},
}

@inproceedings{isokane2018probabilistic,
  title={Probabilistic plant modeling via multi-view image-to-image translation},
  author={Isokane, Takahiro and Okura, Fumio and Ide, Ayaka and Matsushita, Yasuyuki and Yagi, Yasushi},
  booktitle=CVPR,
  pages={2906--2915},
  year={2018}
}

@article{root_recon,
author = {Lu, Yawen and Wang, Yuxing and Chen, Zhanjie and Khan, Awais and Salvaggio, Carl and Lu, Guoyu},
title = {{3D} plant root system reconstruction based on fusion of deep structure-from-motion and {IMU}},
journal = {Multimedia Tools and Applications},
volume = {80},
number = {11},
pages = {17315–17331},
year = {2021},

}

@InProceedings{Yang_2025_ICCV,
    author    = {Yang, Yang and Mao, Dongni and Santo, Hiroaki and Matsushita, Yasuyuki and Okura, Fumio},
    title     = {{NeuraLeaf}: Neural Parametric Leaf Models with Shape and Deformation Disentanglement},
    booktitle = ICCV,
    pages     = {28167-28176},
    year      = {2025},

}

@article{li2025survey,
  title={A survey on {3D} reconstruction techniques in plant phenotyping: from classical methods to neural radiance fields ({NeRF}), {3D} Gaussian splatting ({3DGS}), and beyond},
  author={Li, Jiajia and Qi, Xinda and Nabaei, Seyed Hamidreza and Liu, Meiqi and Chen, Dong and Zhang, Xin and Yin, Xunyuan and Li, Zhaojian},
  journal={arXiv preprint arXiv:2505.00737},
  year={2025}
}

@Article{adtree,
AUTHOR = {Du, Shenglan and Lindenbergh, Roderik and Ledoux, Hugo and Stoter, Jantien and Nan, Liangliang},
TITLE = {{AdTree}: Accurate, Detailed, and Automatic Modelling of Laser-Scanned Trees},
JOURNAL = {Remote Sensing},
VOLUME = {11},
NUMBER = {18},
YEAR = {2019},
}

@article{livnt_tree,
author = {Livny, Yotam and Yan, Feilong and Olson, Matt and Chen, Baoquan and Zhang, Hao and El-Sana, Jihad},
title = {Automatic reconstruction of tree skeletal structures from point clouds},
journal = TOG,
volume = {29},
number = {6},
year = {2010},

}

@inproceedings{liu2025treeformer,
  title={{TreeFormer}: Single-View Plant Skeleton Estimation via Tree-Constrained Graph Generation},
  author={Liu, Xinpeng and Santo, Hiroaki and Toda, Yosuke and Okura, Fumio},
  booktitle=WACV,
  pages={8165--8175},
  year={2025}
}

@Manual{   blender
    ,author = {Blender Online Community},
   title = {Blender - a 3D modelling and rendering package},
   organization = {Blender Foundation},
   address = {Stichting Blender Foundation, Amsterdam},
   year = {2018},
   url = {http://www.blender.org},
 }

\end{document}